\documentclass{article} 
\usepackage{iclr2027_conference,times}

\usepackage[T1]{fontenc}
\usepackage[utf8]{inputenc}
\usepackage{latexsym}
\usepackage{microtype}
\usepackage{inconsolata}
\usepackage{graphicx}
\usepackage{booktabs}
\usepackage{wrapfig}
\usepackage{makecell}
\usepackage{subcaption}
\usepackage{tabularx}
\usepackage{amsmath,amssymb}
\usepackage{multirow}
\usepackage{xcolor}
\usepackage{hyperref}
\usepackage{url}
\newcommand{\ensurefigurespace}[1]{%
  \par\penalty0\begingroup
  \dimen0=\dimexpr\pagegoal-\pagetotal\relax
  \ifdim\dimen0<#1\relax\newpage\fi
  \endgroup
}

\definecolor{darkblue}{rgb}{0, 0, 0.5}
\hypersetup{colorlinks=true, citecolor=darkblue, linkcolor=darkblue, urlcolor=darkblue}

\iclrfinalcopy

\title{Who Is Left of Whom? \\
Tracing Spatial Evidence and Role Binding in Relative-Position Reasoning}

\author{Yingjin Song, Denis Paperno \& Albert Gatt  \\
Utrecht University\\
Utrecht, The Netherlands \\
\texttt{\{y.song5, d.paperno, a.gatt\}@uu.nl}
}

\begin{document}

\maketitle

\begin{abstract}

High instance-level accuracy can mask inconsistencies in spatial reasoning when objects exchange positions or their roles are reversed in the query.
The internal representations supporting relative-position reasoning remain poorly understood. 
We investigate two complementary components of this process: tracking object locations in the input and representing their query roles.
Across three VLMs with visual or textual inputs and their language-model backbones, activation patching reveals a staged progression from early-layer source representations through intermediate-layer query-object representations to late-layer answer states. 
Targeted interventions further establish causal links along this progression: manipulating source-side representations shifts location information at query-object mentions and ultimately alters relation predictions.
Beyond object-location information, we also identify a stable query-side direction associated with the roles of the two objects in the comparison.  
Steering along directions estimated on synthetic scenes generalizes to natural-image benchmarks, improving accuracy and both forms of paired consistency in most settings without retraining.
Our findings reveal complementary components of relational reasoning across visual and textual settings and show how targeted interventions can improve the consistency of models' behavior.

\end{abstract}

\section{Introduction}
Spatial reasoning is a fundamental capability for models operating in visual, linguistic and embodied environments~\citep{du2024embspatial, zhang2026theory, hong2026esibench}. VLMs infer relations from visual layouts and object configurations~\citep{hudson2019gqa, liu-etal-2023-visual, yin2025spatial, Yang_2025_CVPR, cheng2025vstarbenchmarkingvideollmsvideo}, whereas text-only Large Language Models (LLMs) construct spatial structure from linguistic descriptions of object positions, paths, and relations~\citep{DBLP:journals/corr/WestonBCM15, mirzaee2021spartqa, shi2022stepgame, jiang2026spatialtext}. Much of the progress in spatial reasoning is measured through benchmarks, where accuracy is the primary evaluation metric. However, we also need to understand the representations underlying this task, and to assess the extent to which such representations support consistent and robust reasoning.

\begin{figure}[t]
\centering
\includegraphics[width=\textwidth]{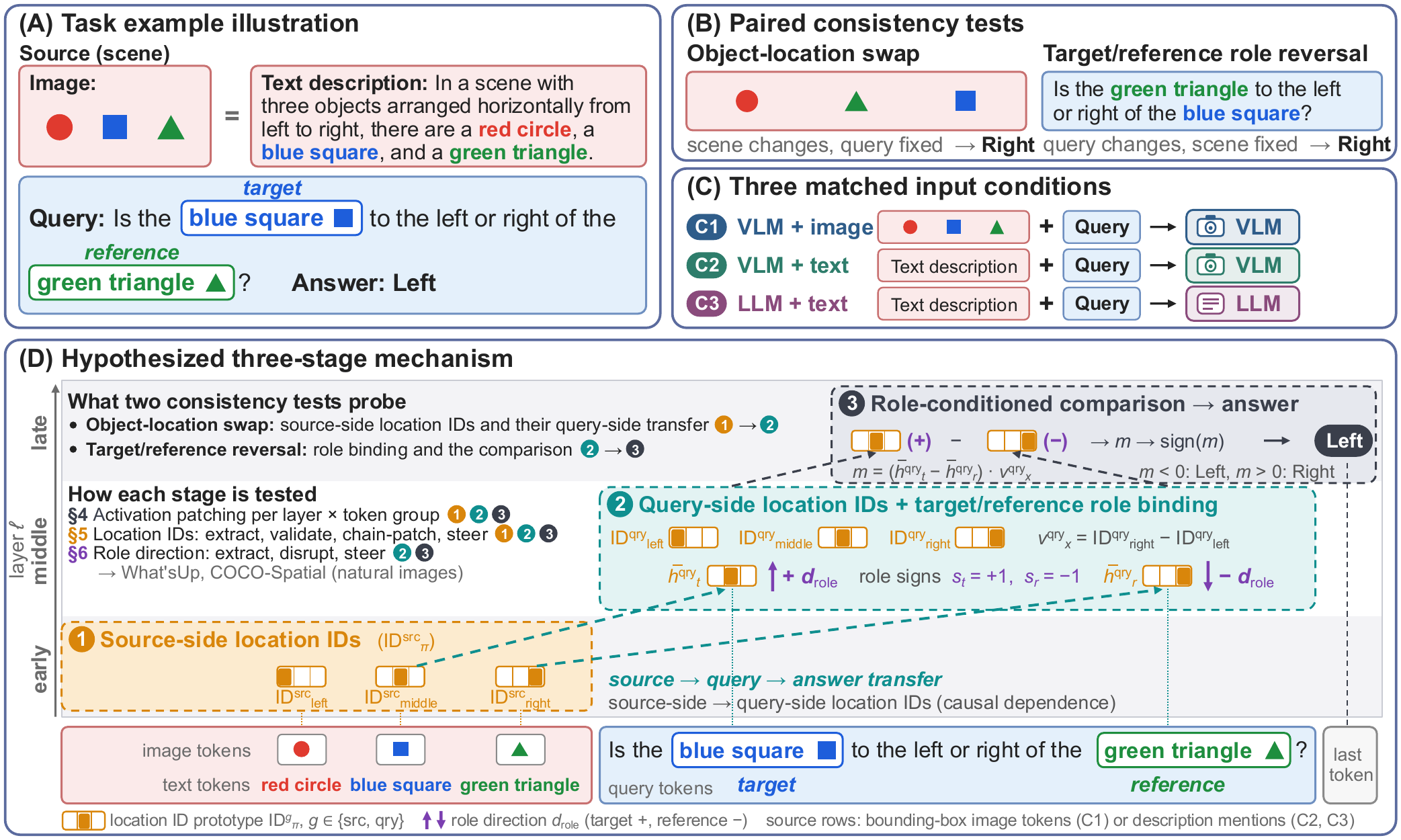}
\caption{\textbf{Overview of the task and mechanistic analysis.}
(A) A scene, given as an image or an equivalent text description (the \emph{source}), and a query about the position of a \emph{target} object relative to a \emph{reference} object.
(B) Object-location swaps and target/reference reversals, both reversing the correct answer.
(C) Matched input conditions: VLM + image, VLM + text and LLM + text.
(D) Hypothesized mechanism: source representations influence query-side location information, while target/reference roles condition its use in relation prediction. We test this account through activation patching and targeted interventions, and evaluate role-direction steering on natural images.}
\label{fig:task_illustration}
\end{figure}

In principle, answering the question ``Is X to the left or right of Y?'' (
\autoref{fig:task_illustration}A) requires at least two computations: (i) recovering the positions of the queried objects and (ii) assigning them to the correct relational roles:  the \emph{target} (object being localized) and \emph{reference} (object with respect to which it is localized).\footnote{The \emph{target} and \emph{reference} correspond to the \emph{figure} and \emph{ground} in linguistic literature~\citep{talmy2003toward}.} 
A model's response should then flip correspondingly under two perturbations 
(\autoref{fig:task_illustration}B): (i) \emph{object-location swap}, where objects exchange positions while the query roles remain fixed; and (ii) \emph{target/reference role reversal}, where the scene remains unchanged but the query roles of target and reference are exchanged. 
A model may answer a single query correctly yet fail under either perturbation. Since each perturbation targets one of the two computations, these paired tests motivate our mechanistic study of how models represent object locations and target/reference roles.
Our starting point is recent work identifying content-independent location components in VLMs: \citet{kang2026linear} show that object-word activations carry spatial IDs whose manipulation shifts spatial beliefs, while \citet{cui2026dual} identify a dominant global signal in visual tokens and a secondary pathway in the language backbone.
We characterize such components using location IDs: position-specific prototypes estimated separately at source-object tokens and at query-object mentions (at query mentions, they correspond to the spatial IDs of \citet{kang2026linear}). 
We extend this line of work in three directions. 
First, prior analyses use image inputs. We test whether location IDs also arise when the scene is described in text, both in the VLM and in its LLM backbone. 
Second, we link the two sites: we hypothesize that location information represented at source tokens becomes available at the corresponding object mentions in the query, and test this source-to-query transfer in visual and textual inputs alike. 
Third, since exchanging target and reference roles reverses the answer for the same locations, we estimate a role direction and test its causal role in relation prediction.


We study three input conditions (\autoref{fig:task_illustration}C): a VLM receiving an image and query (VLM+image) or an equivalent  textual scene description and query (VLM+text), and its LLM backbone receiving the same textual input (LLM+text).
This design allows us to examine differences associated with input modality and multimodal training.
We first evaluate model behavior using paired object-location-swap and target/reference-role-reversal tests. 
As outlined in \autoref{fig:task_illustration}D, we then use activation patching to localize task-relevant information across layers and semantic token groups, extract location IDs, and test their causal effects through source-to-query patching and steering. 
We
evaluate target/reference role directions through both destructive interventions and positive steering. 

\paragraph{Contributions.}
First, through paired object-location-swap and target/reference-reversal consistency tests across the three input settings, we show that high instance-level accuracy does not guarantee relational consistency. 
Second, across visual and textual scene inputs, we characterize location representations at source and query tokens, show that source patching alters query-side location information and answer preferences, and establish that steering query-object states with location-ID differences changes relation predictions.
Third, we identify a stable target/reference role direction, whose disruption degrades role-sensitive predictions. Steering along directions estimated on synthetic scenes improves accuracy and paired consistency on What’sUp \citep{kamath-etal-2023-whats} and COCO-Spatial \citep{lin2014microsoft} in most settings without retraining.

\section{Related work}
\paragraph{Mechanistic interpretability}  studies how neural networks represent information and generate outputs \citep{saphra-wiegreffe-2024-mechanistic}. In language models, probing, logit lens, activation patching, sparse autoencoders and steering have been widely used to localize task-relevant representations and test their causal roles \citep{belinkov2022probing, marks2024the, nostalgebraist2020logitlens, geva-etal-2022-transformer, zhang2024towards, cunningham2023sparse, subramani2022extracting, turner2023steering}. These tools have recently been adapted to VLMs to analyze visual retrieval, cross-modal integration and attention specialization \citep{gandelsman2024interpreting, neo2025towards, ICLR2025_9f14fb9a, krojer2026latentlens, huo-etal-2024-mmneuron, huang2024miner, pach2026sparse, palit2023towards, basu2024understanding, golovanevsky2025vlms, wang-etal-2025-v}. 
Building on these studies, we trace how source-side spatial evidence is propagated to query-side location IDs and how these representations interact with target/reference role binding to support relative-position judgments.

\paragraph{Spatial reasoning} has been widely studied in VLMs using synthetic scenes and human-annotated natural and medical images \citep{kamath-etal-2023-whats, yuksekgonul2023when, liu-etal-2023-visual, Chen_2024_CVPR, ma20253dsrbench, WolDan_Your_MICCAI2025, gholami2026spatial, jia2026omnispatial}, revealing limited robustness and substantial room for improvement. Diagnostic studies attribute these limitations to misallocated visual attention \citep{chen2025spatial}, modality imbalance \citep{qi2025beyond} and the loss of fine-grained spatial details in deep visual features \citep{chen-etal-2025-multimodal}. Closer to our work, mechanistic analyses of image-conditioned VLMs find spatial IDs bound to object words \citep{kang2026linear}, spatial information carried mainly by visual tokens \citep{cui2026dual} and, concurrently, a query-token-mediated pathway for spatial relations \citep{salazar2026pathways}. Spatial reasoning in text-only LLMs has been less explored, with prior work mainly testing whether models infer object positions, paths and relative relations from language \citep{DBLP:journals/corr/WestonBCM15, mirzaee2021spartqa, shi2022stepgame, jiang2026spatialtext, guo2026can}. Our work connects these lines: we trace location representations in VLMs and their LLM backbones under visual and textual input, link source- and query-side location IDs, and identify a separate target/reference role direction.

\section{Task setup and behavioural benchmarking}
\label{sec:benchmarking}

\paragraph{Task setup.}
\label{sec:task_setup}
We study relative-position reasoning with inputs
$x=(S,q(t,r))$, where $S$ is a scene and $q(t,r)$ asks
for the position of a target object $t$ relative to a
reference object $r$ along a specified spatial axis.
Each query concerns two objects in the scene; we also experiment with scenes involving a third, distractor object.
We evaluate the three matched input conditions in
\autoref{fig:task_illustration}C: VLM+image, VLM+text and LLM+text, using
images or textual descriptions of the same scenes.

\paragraph{Paired perturbations.}
We construct two transformations which reverse the ground-truth relation (e.g. from \emph{left} to \emph{right}; cf. \autoref{fig:task_illustration}B) while keeping the queried axis and the order of the answer options fixed.
$
\tau_{\mathrm{loc}}(x)
=
\bigl(S^{t\leftrightarrow r},q(t,r)\bigr)
$
is an \emph{object-location swap}, which exchanges the positions of
the two queried objects while keeping the query fixed. 
In three-object scenes, the distractor
remains unchanged. 
For text inputs, the scene
description is updated to express the swapped layout.
$
\tau_{\mathrm{role}}(x)
=
\bigl(S,q(r,t)\bigr)
$
is a \emph{target/reference role reversal}, where the source
remains unchanged while the two objects exchange
their roles in the query.

\paragraph{Benchmarking setup.}
We evaluate ten VLMs and their LLM backbones on \textbf{Synthetic}, comprising two- and three-object
2D scenes, and \textbf{What'sUp} subsets A and B \citep{kamath-etal-2023-whats}.
We choose What’sUp because its controlled photographs vary spatial relations while preserving object identities, enabling matched location-swap tests on real-world images.
Accuracy is evaluated on original queries, and each paired-consistency
score requires both answers to match their respective ground truths.
Data construction and evaluation details appear in
\autoref{sec:append_data} and \autoref{sec:append_models}.

\begin{figure}[!t]
  \centering
  \includegraphics[width=0.9\textwidth]
  {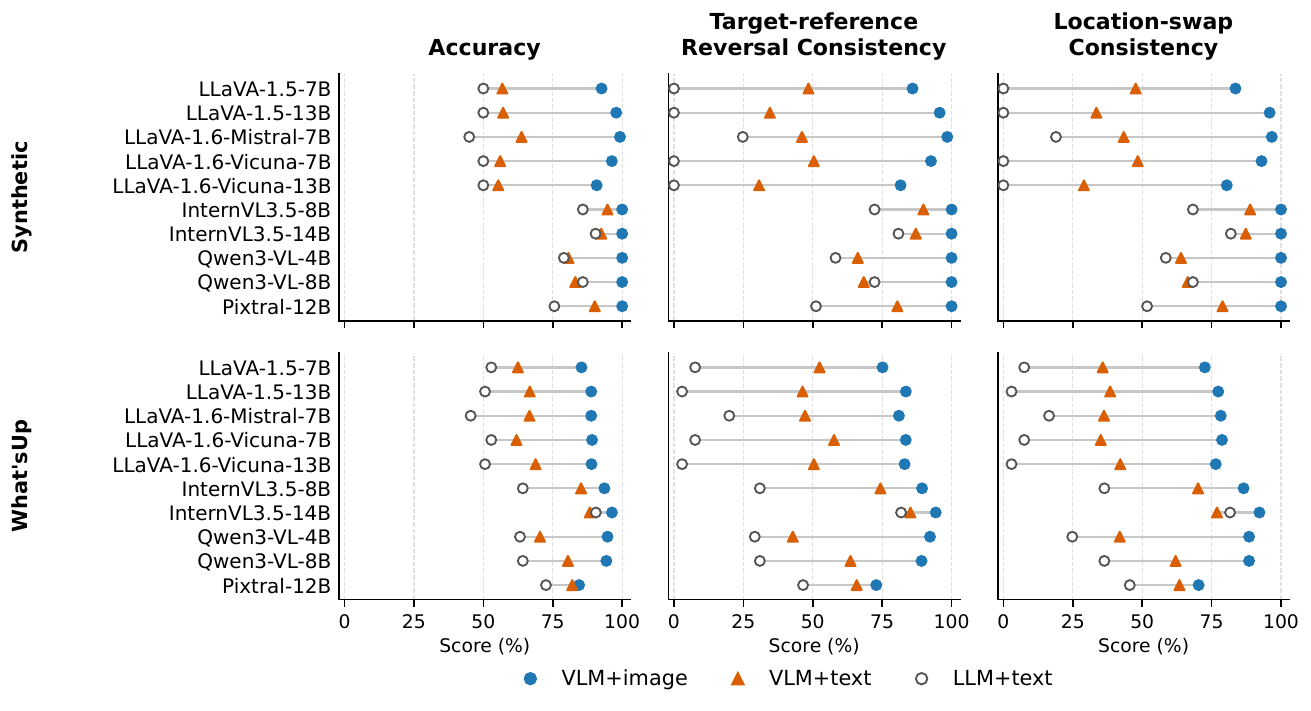}
\caption[Benchmarking results for the three conditions.]{
    Benchmarking results for the three conditions:
    VLM+image, VLM+text and LLM+text.
    Numerical results are provided in
    \autoref{tab:numerical_benchmarking_results}.
}
  \label{fig:benchmarking_results}
\end{figure}

\paragraph{Benchmarking results.}
\autoref{fig:benchmarking_results} shows that VLM+image leads on all three metrics for every model on both benchmarks; five models achieve perfect accuracy and paired consistency on Synthetic. VLM+text generally outperforms LLM+text, with exceptions for Qwen3-VL-8B on all Synthetic metrics and InternVL3.5-14B on What’sUp accuracy and location-swap consistency.
On What’sUp, paired errors remain despite strong original-query accuracy: LLaVA-1.6-Mistral in VLM+image achieves 88.85\% accuracy, compared with 81.03\% target–reference reversal consistency and 78.31\% location-swap consistency. The paired tests also expose fixed-answer behavior: on Synthetic, the Vicuna backbones of LLaVA-1.5 and LLaVA-1.6-Vicuna obtain 50\% accuracy by nearly always answering left or above, yet score 0\% on both consistency metrics. These results motivate examining how models represent and use object locations and query roles across paired inputs.

\section{
Localizing 
source, query and answer-stage representations}
\label{sec:activation_patching}
\paragraph{Why activation patching.}
Building on the behavioral tests in \S\ref{sec:benchmarking}, we use activation patching to localize representations that influence answer preferences under object-location swaps and target/reference role reversals. For each clean--corrupt pair, we replace the residual-stream states of a token group $g$ at layer $\ell$ in the corrupt run with the corresponding clean states and measure recovery of the clean-answer preference~\citep{zhang2024towards}. Comparing the two perturbations 
reveals
where these interventions affect predictions when object locations or query roles change. Our hypothesis (cf. \autoref{fig:task_illustration}D) is that, under location swap, effective patching sites shift with depth from source tokens to query-object mentions and then to the final token. 
We restrict this analysis to pairs in which both runs favor their respective correct answers. This provides a clear clean–corrupt contrast for localizing representations that support correct relative-position judgments.

\paragraph{Experimental Setup.}
We use the three input conditions in \autoref{fig:task_illustration}C. 
Following \S\ref{sec:benchmarking}, we select \texttt{llava-v1.6-mistral-7b} \citep{liu2024llava}, \texttt{InternVL3.5-8B-Instruct} \citep{wang2025internvl3} and \texttt{pixtral-12b} \citep{mistral2024pixtral12b}, and their corresponding LLM backbones: \texttt{Mistral-7B-Instruct} \citep{jiang2023mistral7b}, \texttt{Qwen3-8B} \citep{qwen3technicalreport} and \texttt{Mistral-Nemo-Instruct-2407} \citep{mistral2024nemo}.\footnote{We select these models for mechanistic analyses to cover different model families and performance patterns: 
LLaVA for strong VLM+image but weak VLM+text
and LLM+text performance, InternVL for high Synthetic accuracy across all three conditions, and Pixtral for similar VLM+image
and VLM+text accuracy on What'sUp.
} 
Patched groups follow the three stages: (i) \emph{source} tokens carrying scene evidence
(all visual tokens; the visual tokens inside the bounding boxes of $t$ and $r$; or the
visual tokens in the row/column in which $t/r$ are located; for text inputs, the description
spans naming the objects and those stating their arrangement), (ii) \emph{query} tokens (the
joint target+reference mention span, or the two relation option words); and (iii) the
\emph{last token} before answer generation. V0 denotes the multimodal state before the
first LM layer. \autoref{tab:patching_legend} lists them in full.
We retain clean--corrupt pairs for which the clean run favors $a_{\mathrm{clean}}$, the
corrupt run favors $a_{\mathrm{corrupt}}$, and the margin gap is at least $0.25$. Sample counts before and after screening are reported in Appendix F, together with the inclusion criteria for the experiments in §§5–6.

For each clean--corrupt pair, we cache residual-stream activations from the
forward runs on the clean and corrupt inputs, denoted by \(x_{\mathrm{clean}}\)
and \(x_{\mathrm{corrupt}}\), respectively.
At layer \(\ell\) and semantic token group \(g\), full-vector patching constructs
a patched run \(x_{\mathrm{patch}}^{\mathrm{clean},(\ell,g)}\) by rerunning the
corrupted input with
the corresponding residual-stream activations
from the clean run:
$
H_g^{(\ell)}(x_{\mathrm{patch}}^{\mathrm{clean},(\ell,g)})
\leftarrow
H_g^{(\ell)}(x_{\mathrm{clean}}).
$
For any run \(x\), we define the clean-over-corrupt answer margin as
$
M(x)
=
\mathrm{logit}_{x}(a_{\mathrm{clean}})
-
\mathrm{logit}_{x}(a_{\mathrm{corrupt}}),
$
where \(\mathrm{logit}_{x}(a)\) is the logit assigned to answer \(a\) under run
\(x\). We compute the fraction of examples for which the patched run restores
the preference for the clean answer with \emph{Clean-Answer Recovery Rate}
$
\mathrm{CARR}_{\ell,g}
=
\mathbb{E}\left[
\mathbf{1}\{M(x_{\mathrm{patch}}^{\mathrm{clean},(\ell,g)})>0\}
\right].
$
\footnote{As a complementary continuous metric, we report
$
\mathrm{RestoreScore}_{\ell,g}
=
\mathbb{E}\left[
\frac{
M(x_{\mathrm{patch}}^{\mathrm{clean},(\ell,g)})-M(x_{\mathrm{corrupt}})
}{
M(x_{\mathrm{clean}})-M(x_{\mathrm{corrupt}})
}
\right]
$
in \autoref{append_patching_results},
which measures the fraction of the clean--corrupt margin gap recovered by
patching.}

\begin{figure}[!t]
    \centering
    \includegraphics[width=\textwidth]{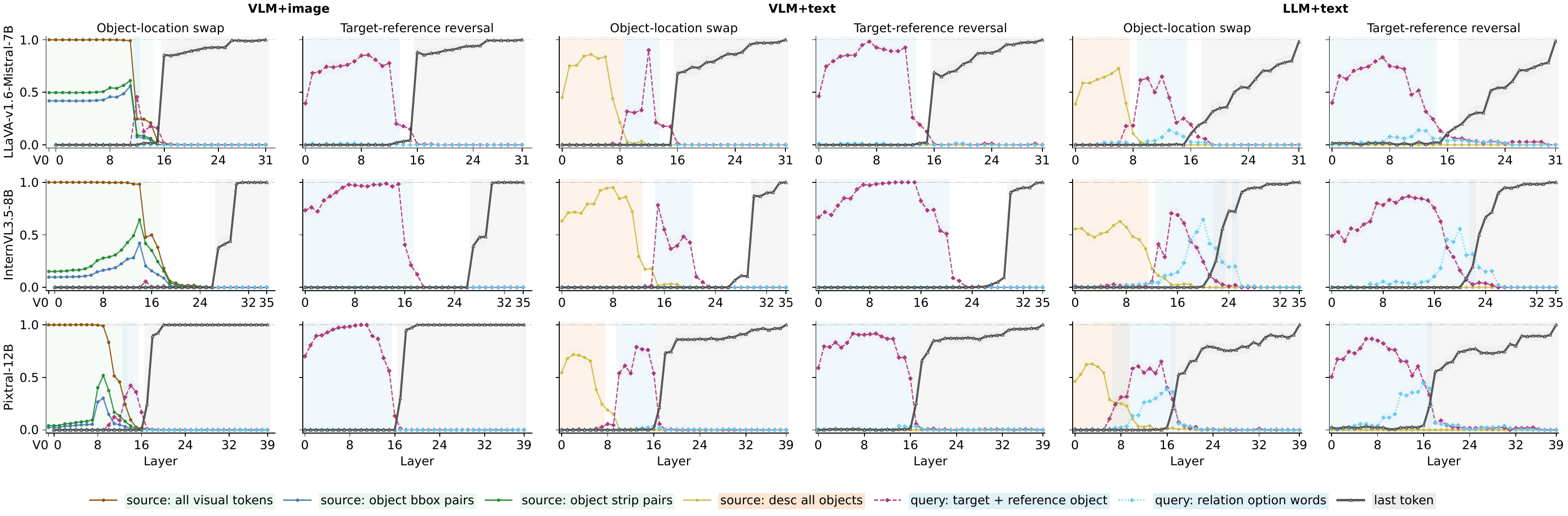}
    \caption{
    Layerwise activation patching clean-answer recovery rate across input conditions, token sources, corruption types and models in the 3-object synthetic dataset. 
    }
    \label{fig:3obj-token-source-restoration}
\end{figure}

\paragraph{Results.}
\autoref{fig:3obj-token-source-restoration} shows layer-dependent patching patterns in the 3-object scene dataset.\footnote{See results for the 2-object scene dataset in \autoref{append_patching_results}.}
Under \emph{location swap}, recovery shifts from source tokens in early-to-middle layers to query-object tokens in middle layers and the final token in late layers.
Under \emph{target-reference reversal}, query-object and late-layer final-token patches dominate recovery.
Source recovery remains near zero as expected, since source tokens precede the query and have identical states across the pair under causal attention.

The most recoverable groups further depend on the input condition.
In VLM+image, location-swap recovery is dominated by visual tokens: patching all visual tokens nearly fully restores the clean answer, while localized visual patches are weaker, with object-strip tokens outperforming object bounding-box tokens.
In the two text-input conditions, recovery shifts to textual source and query groups.
Under location swap, scene-description object mentions are the strongest source-side text group, peaking at about $0.95$ in VLM+text and $0.73$ in LLM+text.
Under target-reference reversal, \textbf{jointly patching the target and reference query spans produces strong clean-answer recovery}.

Comparing VLM+text and LLM+text reveals differences in patching effects across token groups. \textbf{VLM+text generally shows higher recovery from scene-description and query-object patches, whereas patches at relation-option tokens have larger effects in LLM+text}. 
We next examine what location information is represented at source and query-object tokens and whether interventions on these representations affect relation predictions.

\begin{table}[t]
\caption{Effects of source patching on query-side location information and answer preferences. Cells report all-layer mean corrupt-directed score shifts and corrupt-answer flip rates after patching corrupt-run source states into the clean run ($\Delta m^{\rightarrow \mathrm{corr}}$ / flip \%).}
\label{tab:source-to-query-chain}
\centering
\resizebox{\textwidth}{!}{%
\begin{tabular}{llccc}
\toprule
Setting & Patch group & LLaVA $\Delta m$/ flip & InternVL $\Delta m$/ flip & Pixtral $\Delta m$/ flip  \\
\midrule
VLM + image & All visual & 1.718 / 62.9\% & 21.660 / 69.1\% & 3.877 / 51.2\% \\
VLM + image & Object+strip & 0.881 / 34.0\% & 9.691 / 11.6\% & 1.477 / 5.1\% \\
VLM + text & All objects & 0.493 / 17.7\% & 7.912 / 24.3\% & 0.783 / 13.7\% \\
LLM + text & All objects & 0.416 / 20.0\% & 3.586 / 15.6\% & 0.701 / 10.3\% \\
\bottomrule
\end{tabular}
}
\end{table}

\begin{figure}[t]
    \centering
    \includegraphics[width=\textwidth]{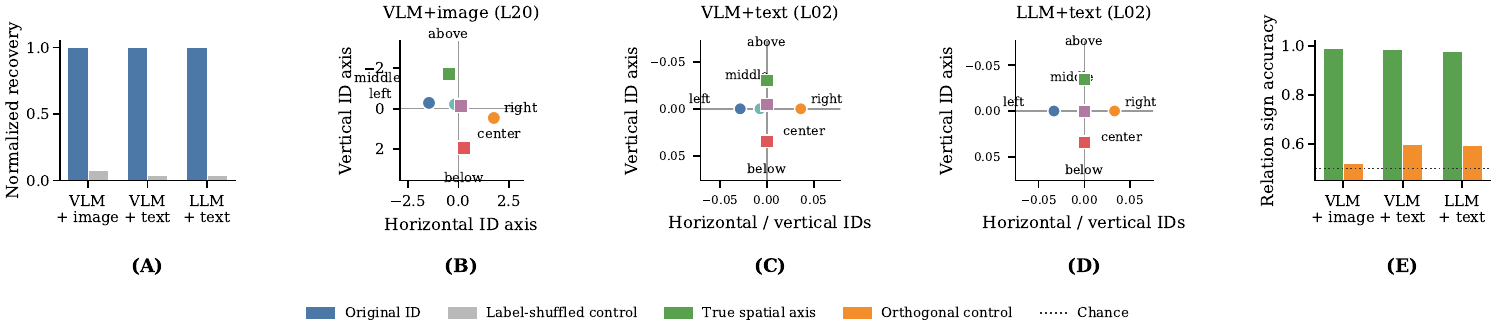}
    \caption{
    Location IDs in LLaVA-1.6-Mistral-7B on three-object scenes: (A) held-out location prediction with shuffled-label controls; (B–D) prototype visualization along axes defined by endpoint-ID contrasts; and (E) query-side relation-sign prediction with orthogonal-axis controls.
    }
    \label{fig:llava_spatial_id_validation}
\end{figure}

\section{Extracting and intervening on object-centered location IDs}
Recent work shows that image-conditioned VLMs encode object locations in object-centered spatial representations, tied to either source-side visual tokens or query-side object tokens~\citep{kang2026linear, cui2026dual}. 
We extend this work
, 
comparing matched image and textual sources, analyzing both VLMs and their LLM backbones, and tracing a causal source-query-prediction pathway. 


\paragraph{Location ID extraction.}
\label{sec:location_ids}
We extract location IDs at two sites, \(g\in\{\mathrm{src},\mathrm{qry}\}\), which are per-object subsets of the source/query token groups in \autoref{tab:patching_legend}.
For example \(i\), object \(o\),  layer \(\ell\), let \(h^{g}_{i,o,\ell}\)
denote the residual-stream state of \(o\) pooled over its tokens at site \(g\):
for \(g=\mathrm{src}\), image tokens inside its bounding box (image inputs)
or the description span naming it (text inputs); for \(g=\mathrm{qry}\), the
query span mentioning it as target or reference.
Each object has an attribute label \(\rho_{i,o}\) (e.g., ``red
circle'', ``blue square'') and a location label \(\pi_{i,o}\): left/right or
above/below in two-object scenes, and left/middle/right or above/center/below in
three-object scenes. 
For image inputs, \(\pi_{i,o}\) is the object's 
location in the visual
scene. 
For text inputs, \(\pi_{i,o}\) denotes the position implied by the description instead of the order of mention. A query mention inherits its source object's label
, so both sites share the label.

To extract location-related components while controlling for object attributes, we center each object state by subtracting the mean activation for objects with the same attribute.
Let \(\mu^{g}_{\rho,\ell}\) denote the mean activation for attribute \(\rho\) at site \(g\) and layer \(\ell\).
We define the centered object state as
$
\bar h^{g}_{i,o,\ell}
=
h^{g}_{i,o,\ell}
-
\mu^{g}_{\rho_{i,o},\ell}.
$
The \emph{location ID} of label $\pi$ at site $g$ and layer $\ell$ is the
prototype obtained by averaging the attribute-centered
states of objects assigned that location label:
$
\mathrm{ID}^{g}_{\pi,\ell}
=
\mathbb{E}\!\left[
\bar h^{g}_{i,o,\ell}
\mid
\pi_{i,o}=\pi
\right].
$
Spatial axes are defined as contrasts between location IDs at opposite ends of each axis, e.g.,
$
v^{g}_{x,\ell}
=
\mathrm{ID}^{g}_{\mathrm{right},\ell}
-
\mathrm{ID}^{g}_{\mathrm{left},\ell},
\qquad
v^{g}_{y,\ell}
=
\mathrm{ID}^{g}_{\mathrm{above},\ell}
-
\mathrm{ID}^{g}_{\mathrm{below},\ell}.
$
Source-side and query-side IDs are estimated independently and need not coincide as vectors; they are linked by the shared label \(\pi\) and, as tested below, by the causal dependence of query-side states on source-side states.

The extracted IDs support held-out location prediction and query-side relational comparison, outperforming the shuffled-label and orthogonal-axis controls, respectively (\autoref{fig:llava_spatial_id_validation}A,E). \autoref{fig:llava_spatial_id_validation}B–D illustrates prototype geometry along axes defined by endpoint-ID contrasts. 
Full results are provided in \autoref{sec:append-spatial-id-validation}.

\paragraph{Source-to-query transfer.}
We test whether changing source representations affects downstream location information at query-object mentions and answer preferences. At layer $\ell$, we replace the activations of source group g in the clean run with the corresponding corrupt-run activations: 
$H_g^{(\ell)}(x_{\mathrm{patch}}^{\mathrm{corrupt},(\ell,g)})
\leftarrow
H_g^{(\ell)}(x_{\mathrm{corrupt}}).$
Let
\(m_i(x;\ell')=\bigl\langle \bar h^{\mathrm{qry}}_{i,t,\ell'}-\bar h^{\mathrm{qry}}_{i,r,\ell'},\, v^{\mathrm{qry}}_{\cdot,\ell'}\bigr\rangle\)
denote the query-side \emph{location-comparison score} for example \(i\): the target-minus-reference difference of the centered query-object states in run \(x\), projected onto the query-side axis of the queried dimension at a downstream layer \(\ell'>\ell\).

We first define the corruption direction as
\(s_i^{\mathrm{corrupt}} = \operatorname{sgn}\!\left(m_i(x_{\mathrm{corrupt}};\ell') - m_i(x_{\mathrm{clean}};\ell')\right)\).
The corrupt-directed query-side location-comparison score shift is then
\(\Delta m_i^{\rightarrow\mathrm{corrupt}} = \left[m_i\!\left(x_{\mathrm{patch}}^{\mathrm{corrupt},(\ell,g)};\ell'\right) - m_i(x_{\mathrm{clean}};\ell')\right]s_i^{\mathrm{corrupt}}\).
Using the clean-over-corrupt margin \(M(x)\), the corrupt-answer flip rate is
$
\mathrm{CAFR}_{\ell,g}
=
\mathbb{E}_i
\left[
\mathbf{1}\{M(x_{\mathrm{patch}}^{\mathrm{corrupt},(\ell,g)})<0\}
\right].
$

Table~\ref{tab:source-to-query-chain} shows that \textbf{patching corrupted source states into clean runs alters query-side location information and answer preferences}. Full visual-token patching yields the largest corrupt-directed score shifts and corrupt-answer flip rates, with weaker effects from localized bbox+strip patches. Description-side object patches also shift scores and induce answer flips in both text-input conditions. We next intervene directly on query-side location representations to test their causal contribution to relation predictions.


\paragraph{Query-side location-ID steering.}
To test whether query-side location information
causally affects relation predictions, we intervene
on the residual-stream states at the target and
reference mentions while keeping the input unchanged.
For each example, let \(\pi_t\) and \(\pi_r\) denote the source-scene spatial labels of the queried target and reference objects on the relevant axis. 
We construct a target-reference spatial-label swap intervention by setting \(\pi'_t=\pi_r\) and \(\pi'_r=\pi_t\), while keeping the target and reference roles unchanged. 
At layer \(\ell\), we add the corresponding query-side ID difference to every token in the query-object span:
$
h_k^{(\ell)}
\leftarrow
h_k^{(\ell)}
+
\alpha
\left(
\mathrm{ID}^{\mathrm{qry}}_{\pi'_o,\ell}
-
\mathrm{ID}^{\mathrm{qry}}_{\pi_o,\ell}
\right),
\qquad
k \in \mathrm{span}(o),\; o \in \{target,reference\}.
$
Here \(k\) indexes token positions, \(\mathrm{span}(o)\) denotes the query-token span of object \(o\), and \(\alpha\) is the steering strength.

\ensurefigurespace{200pt}
\begin{wrapfigure}{r}{0.45\textwidth}
\vspace{-\intextsep}
\captionsetup{font=small}
    \centering
\includegraphics[scale=0.3]{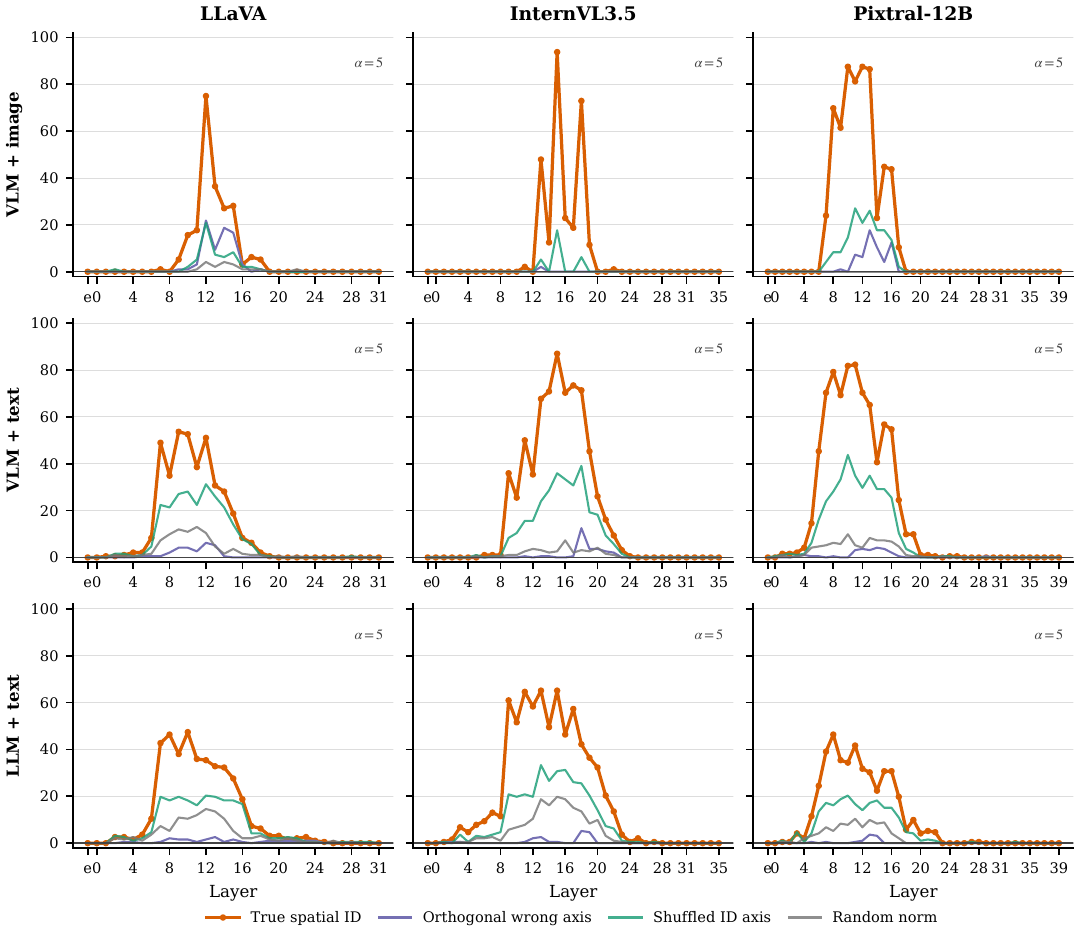}
    \caption{Layer-wise query-side location-ID steering measured by belief-swap rate. }
    \label{fig:query-object-spatial-id-steering}
\end{wrapfigure}

We quantify steering by 
the fraction of examples whose preference changes to the swap-implied answer.
As shown in \autoref{fig:query-object-spatial-id-steering}, {\bf the true query-side ID difference produces the strongest and most consistent belief flips}.
In the VLM+image setting, peak flip rates reach 75.0\%, 93.8\% and 87.5\%, respectively. The effect remains strong in the VLM+text setting 
(53.6\%, 87.0\% and 82.3\%), and persists in the LLM+text setting 
(47.4\%, 65.1\% and 46.4\%). 
The 
flips concentrate in middle layers and are substantially weaker under norm-matched random directions, shuffled-ID directions, and orthogonal controls, indicating that the effect depends on the location-ID alignment of the query-object steering direction.
Query-side location IDs thus specify where each queried object is, but the answer also depends on which object is being localized: the same two location IDs yield opposite answers when the target and reference roles are exchanged.

\par\smallskip\WFclear
\section{Target-reference role direction steering}
\label{sec:role-direction-intervention}


Relative-position 
reasoning additionally depends on assigning the queried objects  target and reference roles, which we call \emph{role binding}.
We test whether 
roles are 
associated with a
stable direction in query-object hidden states and
whether interventions along that direction affect
relation predictions.

\paragraph{Role-contrast geometry.}
We estimate a \emph{role direction} by contrasting the
query-side hidden states of the same object when it
serves as the target vs the reference.
We use target-first (\textbf{TF}) and reference-first (\textbf{RF}) query templates\footnote{
For example, TF: ``Is \underline{the blue square}$_t$ to the left or right of \underline{the green triangle}$_r$?'';
RF: ``Relative to \underline{the green triangle}$_r$, is \underline{the blue square}$_t$ on the left or right?''
In both cases, the blue square is the target and the green triangle is the reference.
}, which place the target and reference in opposite mention orders while preserving the roles and the queried relation.
Let \(\mathcal{C}\) contain training occurrences \((p,o)\) where \(p\) indexes a matched role-reversal pair and object \(o\) appears once as the queried target and once as the queried reference across the two queries in the pair, within each template.
For each \((p,o)\in\mathcal{C}\), we define the joint object-level role contrast as
$
d_{p,o}^{(\ell)}
=
\frac{1}{2}
\sum_{t\in\{\mathrm{TF},\mathrm{RF}\}}
\left(
h_{p,o,t,\mathrm{target}}^{(\ell)}
-
h_{p,o,t,\mathrm{reference}}^{(\ell)}
\right).
$
\ensurefigurespace{180pt}
\begin{wrapfigure}{r}{0.45\textwidth}
\vspace{-\intextsep}
\captionsetup{font=small}
  \centering
  \includegraphics[width=\linewidth]{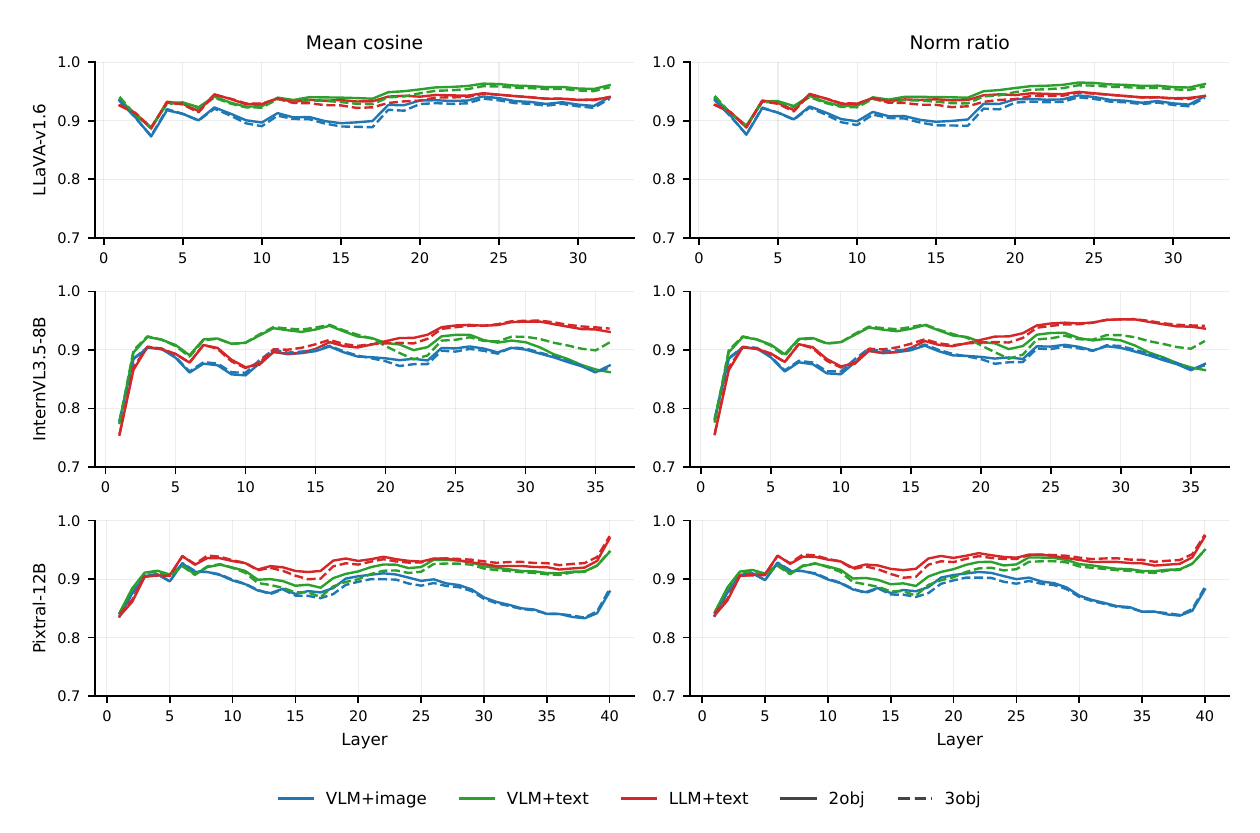}
\caption{
Joint TF/RF role-contrast geometry across layers.
High held-out mean cosine and norm ratio indicate that the joint object-level contrasts share a stable direction.
}
  \label{fig:role_geometry}
\end{wrapfigure}
Here \(h_{p,o,t,\mathrm{target}}^{(\ell)}\) and \(h_{p,o,t,\mathrm{reference}}^{(\ell)}\) are the query-object hidden states of the same object \(o\), pooled over its query-token span, when it is queried as the target and 
as the reference, respectively, under template \(t \in \{TF, RF \}\).
Averaging these joint object-level contrasts gives the joint role direction:
$ d_{\mathrm{role}}^{(\ell)} = \frac{1}{|\mathcal{C}|} \sum_{(p,o)\in \mathcal{C}} d_{p,o}^{(\ell)}. $
We measure the alignment of held-out joint contrasts with the
estimated role direction using the mean cosine
$
\mathbb{E}_{(p,o)\in\mathcal{E}}
\left[
\cos\left(
d_{p,o}^{(\ell)},
d_{\mathrm{role}}^{(\ell)}
\right)
\right],
$
and quantify their concentration using the norm ratio
$
\frac{
\left\|
\mathbb{E}_{(p,o)\in\mathcal{E}}
\left[d_{p,o}^{(\ell)}\right]
\right\|
}{
\mathbb{E}_{(p,o)\in\mathcal{E}}
\left[
\left\|d_{p,o}^{(\ell)}\right\|
\right]
}.
$
A high mean cosine indicates that held-out joint contrasts align
with the estimated role direction, while a high norm ratio indicates
that averaging preserves a large fraction of their mean individual
norm.
Both metrics remain high across models and conditions in \autoref{fig:role_geometry}, suggesting that {\bf the joint target/reference role contrast forms a stable hidden-state direction}.

\ensurefigurespace{100pt}
\begin{wrapfigure}{r}{0.45\textwidth}
    \vspace{-\intextsep}
    \centering
    \captionsetup{font=small,skip=4pt}

    \includegraphics[width=\linewidth]
    {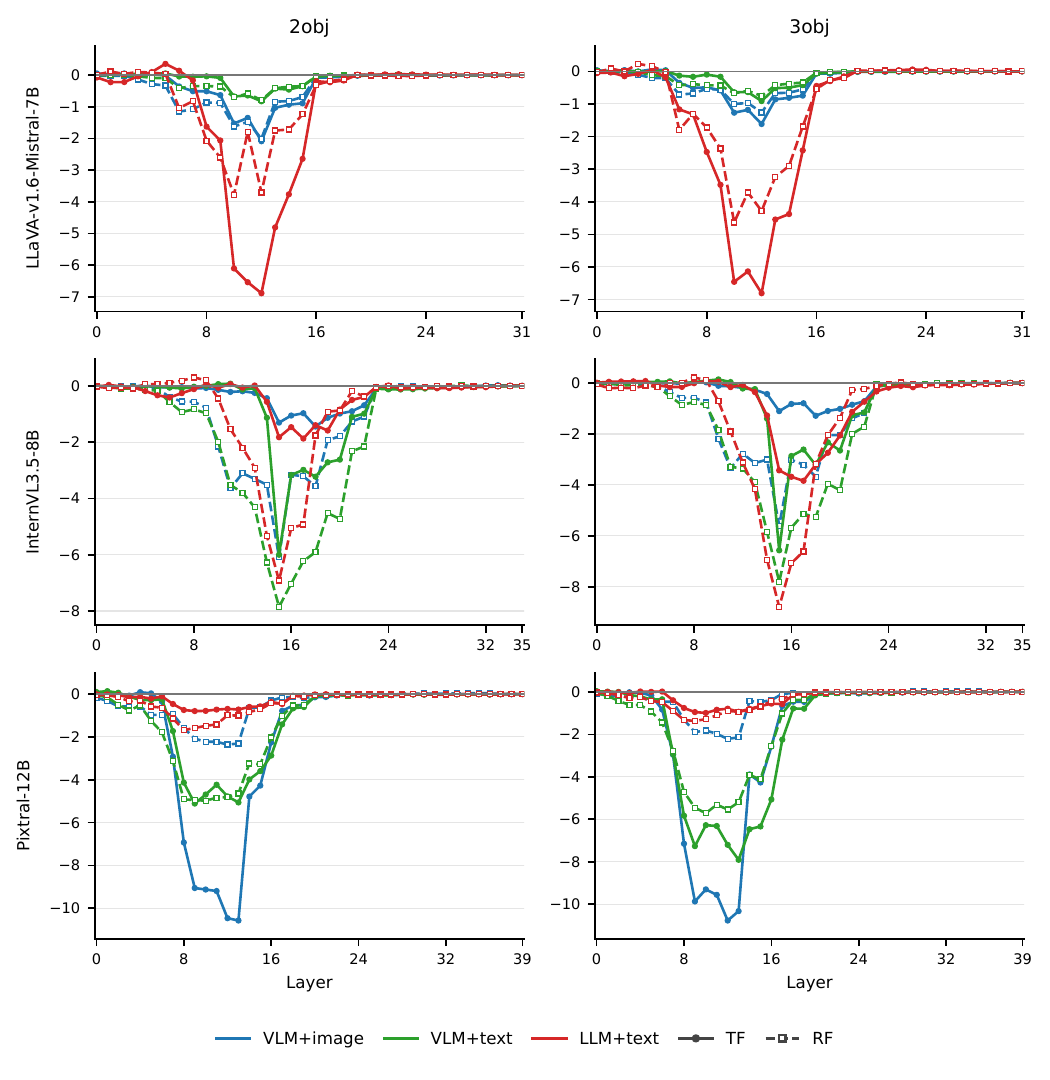}

    \caption{
    Layer-wise joint role-direction intervention at \(\alpha=1\).
    Curves show the mean gap
    \(\Delta M_{\mathrm{role}}-\Delta M_{\mathrm{orth}}\). Negative values indicate stronger disruption
along the joint role direction.}
    \label{fig:role_direction_layers}
\end{wrapfigure}
\paragraph{Role-direction intervention.}
We test the causal contribution of the 
role direction
by applying destructive updates to all tokens in the
query-object spans at a single 
layer:
$
h_{\mathrm{target}}^{(\ell)}
\leftarrow h_{\mathrm{target}}^{(\ell)}
-\alpha d_{\mathrm{role}}^{(\ell)},
\qquad
h_{\mathrm{reference}}^{(\ell)}
\leftarrow h_{\mathrm{reference}}^{(\ell)}
+\alpha d_{\mathrm{role}}^{(\ell)}.
$
For each setting, we evaluate 128 held-out scene groups
under both TF and RF templates.
The full-layer sweep uses the unnormalized direction
at the prespecified strength \(\alpha=1\), corresponding
to one estimated role-contrast vector.
We also compare \(\alpha\in\{0.5,1,1.5\}\), averaging
separate single-layer interventions within fixed layer bands.

We measure the intervention-induced change \(\Delta M\)
in the candidate-answer margin
\(M(x)=\mathrm{score}_{x}(a^+)-\mathrm{score}_{x}(a^-)\),
where \(a^+\) and \(a^-\) are the correct and role-reversed
answers, and report
$
D=\Delta M_{\mathrm{role}}-\Delta M_{\mathrm{orth}}.
$
Here, \(\Delta M_{\mathrm{orth}}\) averages five separately
evaluated, norm-matched orthogonal random controls;
negative \(D\) indicates stronger disruption along
the joint direction.
The strongest negative gaps concentrate in early-to-middle
layers (\autoref{fig:role_direction_layers}).
Within the fixed bands, the plotted gaps become more
negative with increasing \(\alpha\) under both TF and RF
templates (\autoref{fig:role_direction_dose}).
These effects support a strength-dependent causal
contribution of the joint role-related direction
under both mention orders.




\begin{table}[t]
\centering
\small
\caption{Activation steering on What’sUp-A using the joint target/reference role direction estimated on synthetic data. Bold indicates improvements with 95\% confidence intervals excluding zero.}
\label{tab:steered_generation_whatsup}
\resizebox{\textwidth}{!}{%
\begin{tabular}{ccccc}
\toprule
Model & Condition & Accuracy &
\makecell{Target-Reference\\Reversal Consistency} &
\makecell{Location-Swap\\Consistency} \\
\midrule
LLaVA-1.6 & VLM+image & $82.2 \rightarrow 84.0$ {\scriptsize $\mathbf{(+1.8)}$} & $70.6 \rightarrow 73.9$ {\scriptsize $\mathbf{(+3.3)}$} & $66.3 \rightarrow 67.5$ {\scriptsize $(+1.3)$} \\
LLaVA-1.6 & VLM+text & $66.3 \rightarrow 65.1$ {\scriptsize $(-1.2)$} & $52.2 \rightarrow 54.9$ {\scriptsize $\mathbf{(+2.7)}$} & $39.7 \rightarrow 39.2$ {\scriptsize $(-0.5)$} \\
LLaVA-1.6 & LLM+text & $63.8 \rightarrow 67.1$ {\scriptsize $\mathbf{(+3.3)}$} & $38.6 \rightarrow 43.9$ {\scriptsize $\mathbf{(+5.3)}$} & $33.5 \rightarrow 38.2$ {\scriptsize $\mathbf{(+4.7)}$} \\
\midrule
InternVL3.5 & VLM+image & $97.3 \rightarrow 97.4$ {\scriptsize $(+0.1)$} & $96.1 \rightarrow 96.6$ {\scriptsize $(+0.5)$} & $94.2 \rightarrow 94.2$ {\scriptsize $(+0.0)$} \\
InternVL3.5 & VLM+text & $87.1 \rightarrow 89.7$ {\scriptsize $\mathbf{(+2.6)}$} & $78.7 \rightarrow 80.9$ {\scriptsize $\mathbf{(+2.2)}$} & $74.1 \rightarrow 77.8$ {\scriptsize $\mathbf{(+3.7)}$} \\
InternVL3.5 & LLM+text & $70.8 \rightarrow 73.2$ {\scriptsize $\mathbf{(+2.4)}$} & $46.9 \rightarrow 50.3$ {\scriptsize $\mathbf{(+3.3)}$} & $45.3 \rightarrow 49.7$ {\scriptsize $\mathbf{(+4.4)}$} \\
\midrule
Pixtral & VLM+image & $83.5 \rightarrow 85.0$ {\scriptsize $\mathbf{(+1.6)}$} & $69.1 \rightarrow 72.3$ {\scriptsize $\mathbf{(+3.2)}$} & $67.3 \rightarrow 70.5$ {\scriptsize $\mathbf{(+3.2)}$} \\
Pixtral & VLM+text & $77.0 \rightarrow 81.9$ {\scriptsize $\mathbf{(+4.9)}$} & $57.8 \rightarrow 65.7$ {\scriptsize $\mathbf{(+7.8)}$} & $54.7 \rightarrow 64.1$ {\scriptsize $\mathbf{(+9.4)}$} \\
Pixtral & LLM+text & $74.5 \rightarrow 78.7$ {\scriptsize $\mathbf{(+4.2)}$} & $50.0 \rightarrow 58.8$ {\scriptsize $\mathbf{(+8.8)}$} & $49.6 \rightarrow 58.0$ {\scriptsize $\mathbf{(+8.3)}$} \\
\bottomrule
\end{tabular}
}
\end{table}

\par\smallskip\WFclear
\paragraph{Role-direction steering.}
We test whether amplifying the query-specified role signal
along the joint direction estimated on synthetic two-object
data improves spatial reasoning on What'sUp-A and COCO-spatial.\footnote{
We use What'sUp subset A and COCO-Spatial because, like our synthetic data,
both query only \textit{left}, \textit{right}, \textit{above} and \textit{below},
so synthetic-estimated directions apply without changing the answer space.
What'sUp-B involves \textit{in front of}/\textit{behind} and is included in
\autoref{fig:benchmarking_results} but not here.}
We apply the unnormalized direction to query-object tokens:
$
h_k^{(\ell)}
\leftarrow
h_k^{(\ell)}
+\alpha s_o d_{\mathrm{role}}^{(\ell)},
\qquad
k\in\mathrm{span}(o),
$
where \(s_o=+1\) for the target and \(-1\) for the reference.
We retain the layers previously selected for the joint direction
on synthetic validation data and select
\(\alpha\in\{0,0.25,0.5,1,1.5\}\) for the joint direction
on 128 disjoint synthetic two-object validation scene groups.
Selection maximizes original-query generation accuracy,
averaged equally over TF/RF templates and scene-description
variants.
The resulting layer/strength configurations
(\autoref{tab:steering_layer_alpha}) are frozen before
What'sUp and COCO-Spatial evaluation and model parameters remain unchanged.

\autoref{tab:steered_generation_whatsup} reports results pooled over TF and RF queries on What’sUp-A. Joint role-direction steering improves target–reference reversal consistency in all nine settings, with 95\% confidence intervals excluding zero in eight. 
Accuracy and location-swap consistency also improve in most settings. 
The largest gains occur in Pixtral’s text-input conditions: VLM+text gains 4.9\% in accuracy, 7.8\% in reversal consistency, and 9.4\% in location-swap consistency. 
LLaVA-1.6 VLM+text shows a more selective effect, improving reversal consistency despite small declines in the other metrics. 
Applying the same directions, layers and strengths to COCO-Spatial without retuning also improves reversal consistency in all nine settings and accuracy in most (\autoref{tab:steered_generation_coco_spatial}). 
Together, these results show that \textbf{directions estimated on synthetic scenes can improve accuracy and paired consistency on natural-scene benchmarks without retraining}.

\section{Conclusion}
We study the internal mechanisms of relative-position reasoning across matched VLM+image, VLM+text and LLM+text settings. Our behavioral results show that high instance-level accuracy can coexist with inconsistent predictions under object-location swaps and target/reference role reversals. Mechanistically, our analyses support an account in which models compare location information at query-object mentions according to the objects’ target/reference roles. Activation patching reveals a staged progression from early-layer source representations through intermediate-layer query-object representations to late-layer answer states. Source-side interventions alter query-side location information and shift answer preferences, while steering query-object states with location-ID differences changes relation predictions. We further identify a stable query-side direction associated with target/reference roles, whose disruption reduces correct-answer preference relative to matched orthogonal controls. Together, these results establish the causal relevance of both object-location information and query-role representations, supporting their complementary contributions to relative-position reasoning.

These findings extend evidence for causally relevant object-location representations from image-conditioned VLMs to textual scene inputs in both VLMs and their LLM backbones. 
Beyond the controlled mechanistic analyses, steering along role directions estimated on synthetic scenes improves accuracy and paired consistency on natural-image benchmarks in most settings without retraining. 
A broader lesson is that paired behavioral tests become more informative when combined with mechanistic interventions that clarify how models represent and use spatial information, ultimately guiding future efforts to improve the consistency and robustness of relational reasoning.

\section*{AI use statement}
Generative AI tools were used to support code development and debugging, identify relevant literature and assist with manuscript writing and revision, particularly to improve wording and clarity. 
All AI-assisted code, references and text were reviewed and verified by the authors. 
The authors take full responsibility for the final content of this paper.

\section*{Reproducibility Statement}
Data construction procedures and dataset statistics are provided in
\autoref{sec:append_data}, and model checkpoints and benchmark composition
are listed in \autoref{sec:append_models}.
The corresponding experimental sections specify the activation-patching
and steering procedures and the evaluation metrics.
\autoref{app:compute} documents the software environment, numerical
precision, decoding settings, random-seed controls, and compute resources.
Code, experiment configurations and synthetic datasets will be released
upon publication.

\section*{Ethics statement}
\label{sec:append_ethics} 
This work studies spatial reasoning mechanisms in VLMs and LLMs using controlled synthetic data and existing benchmarks. 
We do not collect human-subject data, personal information or sensitive attributes. 
Our goal is to understand the internal mechanisms underlying spatially consistent predictions, which matter for downstream systems such as robotics or assistive technologies.
Activation steering could be misused to manipulate model behavior, but we use it only as an analysis tool for understanding representations and robustness.

\section*{Limitations}
\label{sec:append_limitations}
To enable matched consistency tests and precise causal interventions, we study relative-position judgments (i.e., \textit{left}, \textit{right}, \textit{above}, \textit{below}) in controlled two- and three-object scenes, with objects at fixed grid positions and templated queries. 
The role direction estimated on these scenes nevertheless transfers to natural-image benchmarks without retraining. 
Extending the analysis to depth, distance and multi-object relations, freer layouts and open-ended phrasings is a natural next step. 
Following standard practice for activation patching, our localization analyses use correctly answered pairs and thus characterize how correct judgments are implemented.
Steering on full benchmarks already links the role direction to naturally occurring errors, and attributing individual errors to specific components is left for future work. Our mechanistic analyses cover three open 7B–12B models from different families; larger, MoE and proprietary models remain to be examined.
Finally, role-direction steering offers an initial analysis-oriented intervention that improves paired consistency without retraining. 
Future work can evaluate whether related interventions can be turned into more general post-training methods and whether they transfer to broader capabilities and out-of-distribution spatial tasks.

\bibliographystyle{iclr2027_conference}
\bibliography{references}

\clearpage
\appendix

\begin{figure}[t]
  \centering
  \includegraphics[width=\textwidth]{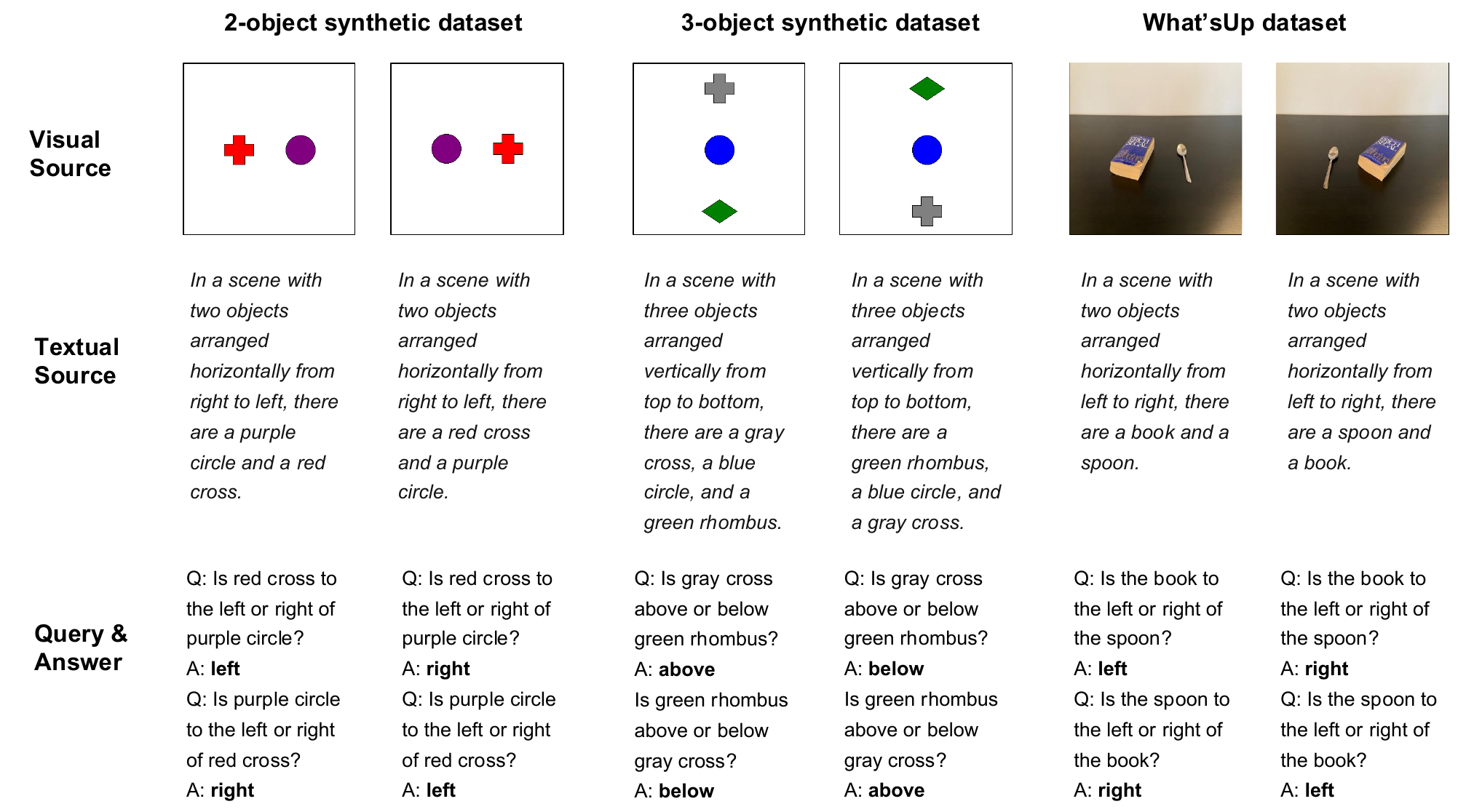}
  \caption{Example data used in our experiments }
  \label{fig:data_illu}
\end{figure}
 
\section{Details of data construction}
\label{sec:append_data}

\autoref{fig:data_illu} shows representative examples from synthetic and natural-image datasets under visual and textual source conditions. For each scene, we construct reciprocal questions by swapping the target and reference objects, enabling analysis of whether models consistently update their answers when the underlying spatial relation is reversed.

We construct two controlled synthetic spatial-reasoning datasets: a two-object dataset and a three-object dataset. Each image is rendered on a 336 \ensuremath{\times} 336 white canvas, with objects placed at fixed anchor points on a 3 \ensuremath{\times} 3 grid. Objects are defined as color-shape conjunctions. We use six colors, red, blue, green, yellow, purple, and gray, and six shapes, circle, square, triangle, cross, ellipse, and rhombus, yielding 36 distinct object types.

The three-object dataset is our primary setting. Each image contains three collinearly arranged objects placed on one of the three rows or one of the three columns. For each three-object scene, we generate six directed pairwise questions, covering all ordered relations among the three objects. Each question mentions only the queried target object and reference object, while the third object remains an unmentioned spectator. The model is required to answer with exactly one word from above, below, left, and right. We balance relation labels, horizontal versus vertical axes, and row/column layouts. The statistics of the synthetic datasets are displayed in \autoref{tab:mechanistic_dataset_statistics}.

Textual descriptions are generated by deterministic rules. For each scene, we generate a description of the object arrangement, such as ``In a scene with three objects arranged horizontally from left to right, there are a red circle, a blue square, and a green triangle.'' We further generate two ordered scene descriptions that list the objects along the queried spatial axis in opposite directions: top-to-bottom and bottom-to-top for vertical layouts, and left-to-right and right-to-left for horizontal layouts. These textual descriptions preserve object identity and spatial structure while expressing scene information linguistically, enabling text-only and cross-modal control conditions.

\begin{table}[t]
\caption{
Statistics of the synthetic datasets used for mechanistic
analyses.
A scene denotes a unique rendered image, and a question denotes one
directed target--reference query.
``Each relation'' reports the number of questions for each of
\{\textit{left}, \textit{right}, \textit{above}, \textit{below}\}.
Each two-object scene contains two reciprocal questions and therefore
one target--reference pair.
Each three-object scene contains all six directed questions among its
three objects, corresponding to three reciprocal target--reference
pairs.
}
\label{tab:mechanistic_dataset_statistics}
\centering
\small
\setlength{\tabcolsep}{3pt}
\resizebox{\textwidth}{!}{%
\begin{tabular}{llrrrrrr}
\toprule
Dataset
& Split
& Scenes
& Questions
& Horizontal
& Vertical
& \makecell{Each\\relation}
& \makecell{Reciprocal\\pairs} \\
\midrule
\multirow{4}{*}{\makecell[l]{Synthetic\\2-object}}
& Train      & 1,512 & 3,024 & 1,512 & 1,512 & 756 & 1,512 \\
& Validation &   324 &   648 &   324 &   324 & 162 &   324 \\
& Test       &   648 & 1,296 &   648 &   648 & 324 &   648 \\
& Total      & 2,484 & 4,968 & 2,484 & 2,484 & 1,242 & 2,484 \\
\midrule
\multirow{4}{*}{\makecell[l]{Synthetic\\3-object}}
& Train      &   504 & 3,024 & 1,512 & 1,512 & 756 & 1,512 \\
& Validation &   108 &   648 &   324 &   324 & 162 &   324 \\
& Test       &   216 & 1,296 &   648 &   648 & 324 &   648 \\
& Total      &   828 & 4,968 & 2,484 & 2,484 & 1,242 & 2,484 \\
\bottomrule
\end{tabular}%
}

\end{table}


\section{Benchmarking models, data and numerical results}
\label{sec:append_models}

The details of the VLMs and their corresponding LLM backbones are displayed in \autoref{tab:benchmark_model_architectures}.

\begin{table}[t]
\caption{Architectures of benchmarked models. Layers denotes the number of transformer blocks in the language backbone.}
\label{tab:benchmark_model_architectures}
\centering
\small
\begin{tabular}{@{}lllr@{}}
\toprule
VLM & Arch. type & LLM backbone & Layers \\
\midrule
LLaVA-v1.5-7B & Projector-concat & Vicuna v1.5 (7B) & 32 \\
LLaVA-v1.5-13B & Projector-concat & Vicuna v1.5 (13B) & 40 \\
LLaVA-v1.6-Mistral-7B & Projector-concat & Mistral (7B) & 32 \\
LLaVA-v1.6-Vicuna-7B & Projector-concat & Vicuna v1.5 (7B) & 32 \\
LLaVA-v1.6-Vicuna-13B & Projector-concat & Vicuna v1.5 (13B) & 40 \\
Pixtral-12B & Projector-concat & Mistral (12B) & 40 \\
Qwen3-VL-4B & Embed-concat & Qwen3 (4B) & 36 \\
Qwen3-VL-8B & Embed-concat & Qwen3 (8B) & 36 \\
InternVL3.5-8B & ViT-MLP-LLM & Qwen3 (8B) & 36 \\
InternVL3.5-14B & ViT-MLP-LLM & Qwen3 (14B) & 40 \\
\bottomrule
\end{tabular}
\end{table}

\autoref{tab:behavior_dataset_statistics} reports TF evaluation counts.
Original images denotes distinct underlying scene images, including
those represented by descriptions in the text conditions.
A scene family groups images linked by location swaps and their
query and description variants.
Accuracy uses original queries; reversal consistency uses pairs
of original queries; location-swap consistency pairs each original
query with its swapped counterpart.
Synthetic text conditions include two description variants per scene.

\begin{table}[t]
\caption{Evaluation counts for \autoref{fig:benchmarking_results} and
\autoref{tab:numerical_benchmarking_results}. Counts are per evaluated model;
text conditions pool the available description variants.}
\label{tab:behavior_dataset_statistics}
\centering
\small
\setlength{\tabcolsep}{3pt}
\resizebox{\textwidth}{!}{%
\begin{tabular}{llrrrrr}
\toprule
Dataset & Condition
& \makecell{Original\\images}
& \makecell{Scene\\families}
& \makecell{Original\\queries}
& \makecell{Reversal\\pairs}
& \makecell{Swap\\anchors} \\
\midrule
Synthetic 2-object & VLM+image & 128 & 128 & 256 & 128 & 256 \\
                  & VLM+text  & 128 & 128 & 512 & 256 & 512 \\
                  & LLM+text  & 128 & 128 & 512 & 256 & 512 \\
\midrule
Synthetic 3-object & VLM+image & 128 & 128 & 256 & 128 & 256 \\
                  & VLM+text  & 128 & 128 & 512 & 256 & 512 \\
                  & LLM+text  & 128 & 128 & 512 & 256 & 512 \\
\midrule
What'sUp-A & VLM+image & 180 & 132 & 360 & 180 & 360 \\
           & VLM+text  & 180 & 136 & 360 & 180 & 360 \\
           & LLM+text  & 180 & 134 & 360 & 180 & 360 \\
\midrule
What'sUp-B & VLM+image & 204 & 154 & 408 & 204 & 408 \\
           & VLM+text  & 204 & 151 & 408 & 204 & 408 \\
           & LLM+text  & 204 & 155 & 408 & 204 & 408 \\
\bottomrule
\end{tabular}%
}
\end{table}

The equivalent numerical results of \autoref{fig:benchmarking_results} are displayed in \autoref{tab:numerical_benchmarking_results}.

\begin{table}[!ht]
\caption{Numerical benchmarking results for VLM+image, VLM+text, and the corresponding LLM+text backbone. Synthetic includes the two- and three-object 2D sets, and What'sUp includes subsets A and B.  }
\label{tab:numerical_benchmarking_results}
\centering
\small
\setlength{\tabcolsep}{3pt}
\resizebox{\textwidth}{!}{%
\begin{tabular}{lrrrrrrrrr}
\toprule
Model
& \multicolumn{3}{c}{Accuracy}
& \multicolumn{3}{c}{\makecell{Target--reference\\reversal consistency}}
& \multicolumn{3}{c}{\makecell{Location-swap\\consistency}} \\
& \makecell{VLM+\\image}
& \makecell{VLM+\\text}
& \makecell{LLM+\\text}
& \makecell{VLM+\\image}
& \makecell{VLM+\\text}
& \makecell{LLM+\\text}
& \makecell{VLM+\\image}
& \makecell{VLM+\\text}
& \makecell{LLM+\\text} \\
\cmidrule(lr){2-4}
\cmidrule(lr){5-7}
\cmidrule(lr){8-10}

\multicolumn{10}{l}{\textit{(a) Synthetic}} \\
LLaVA-1.5-7B         & 92.58  & 56.84 & 50.00 & 85.94  & 48.44 & 0.00  & 83.59  & 47.66 & 0.00  \\
LLaVA-1.5-13B        & 97.85  & 57.13 & 50.00 & 95.70  & 34.57 & 0.00  & 95.90  & 33.50 & 0.00  \\
LLaVA-1.6-Mistral-7B & 99.22  & 63.77 & 44.92 & 98.44  & 46.09 & 24.80 & 96.68  & 43.36 & 18.95 \\
LLaVA-1.6-Vicuna-7B  & 96.29  & 56.05 & 50.00 & 92.58  & 50.39 & 0.00  & 92.97  & 48.44 & 0.00  \\
LLaVA-1.6-Vicuna-13B & 90.82  & 55.37 & 50.00 & 81.64  & 30.66 & 0.00  & 80.47  & 29.00 & 0.00  \\
InternVL3.5-8B       & 100.00 & 94.73 & 85.84 & 100.00 & 89.84 & 72.27 & 100.00 & 88.87 & 68.26 \\
InternVL3.5-14B      & 100.00 & 92.48 & 90.43 & 100.00 & 87.11 & 80.86 & 100.00 & 87.30 & 81.93 \\
Qwen3-VL-4B         & 100.00 & 80.66 & 79.00 & 100.00 & 66.21 & 58.20 & 100.00 & 63.96 & 58.50 \\
Qwen3-VL-8B         & 100.00 & 83.11 & 85.84 & 100.00 & 68.36 & 72.27 & 100.00 & 66.41 & 68.26 \\
Pixtral-12B         & 100.00 & 90.14 & 75.59 & 100.00 & 80.47 & 51.17 & 100.00 & 78.91 & 51.76 \\

\midrule
\multicolumn{10}{l}{\textit{(b) What'sUp}} \\
LLaVA-1.5-7B         & 85.35 & 62.46 & 52.86 & 75.15 & 52.43 & 7.63  & 72.57 & 35.77 & 7.51  \\
LLaVA-1.5-13B        & 88.85 & 66.73 & 50.61 & 83.53 & 46.34 & 2.94  & 77.39 & 38.45 & 2.94  \\
LLaVA-1.6-Mistral-7B & 88.85 & 66.63 & 45.43 & 81.03 & 47.21 & 19.92 & 78.31 & 36.24 & 16.45 \\
LLaVA-1.6-Vicuna-7B  & 89.15 & 61.95 & 52.86 & 83.51 & 57.66 & 7.63  & 78.75 & 35.07 & 7.51  \\
LLaVA-1.6-Vicuna-13B & 88.97 & 68.84 & 50.61 & 83.09 & 50.39 & 2.94  & 76.50 & 42.09 & 2.94  \\
InternVL3.5-8B       & 93.60 & 85.18 & 64.22 & 89.36 & 74.36 & 30.95 & 86.53 & 70.11 & 36.36 \\
InternVL3.5-14B      & 96.32 & 88.33 & 90.52 & 94.30 & 85.18 & 81.81 & 92.23 & 76.95 & 81.65 \\
Qwen3-VL-4B         & 94.72 & 70.39 & 63.20 & 92.22 & 42.81 & 29.12 & 88.49 & 41.90 & 24.81 \\
Qwen3-VL-8B         & 94.31 & 80.48 & 64.22 & 89.17 & 63.58 & 30.95 & 88.49 & 62.04 & 36.36 \\
Pixtral-12B         & 84.49 & 81.99 & 72.56 & 72.88 & 65.82 & 46.50 & 70.33 & 63.43 & 45.53 \\
\bottomrule
\end{tabular}%
}
\vspace{4pt}
\begin{minipage}{\textwidth}
\footnotesize
VLM+text includes a white blank image.
Synthetic uses 256 scene families; What'sUp uses 286/287/289
for VLM+image/VLM+text/LLM+text.
Models sharing a language backbone share LLM+text results.
\end{minipage}
\end{table}

\section{Additional activation patching results}
\label{append_patching_results}

\begin{table}[t]\centering\small
\caption{Token groups patched in all experiments in \autoref{sec:activation_patching}. $t$ means target object and $r$ means reference object.}
\label{tab:patching_legend}
\begin{tabular}{@{}llp{0.5\linewidth}@{}}
\toprule
Stage & Legend label & Tokens \\
\midrule
source & all visual tokens   & every image token (C1 only) \\
source & object bbox pairs   & image tokens inside the bounding boxes of $t$ and $r$ \\
source & object strip pairs  & image tokens in the row (or column) strips through $t$ and $r$ \\
source & desc objects        & description spans naming $t$ and $r$ (C2, C3) \\
source & desc locations      & description spans stating the arrangement (C2, C3) \\
query  & target + reference object & the query spans mentioning $t$ and $r$, patched jointly \\
query  & relation option words     & the two relation words offered in the query \\
final     & last token          & the final input position before answer generation \\
\bottomrule
\end{tabular}
\end{table}

\begin{figure}[t]
    \centering
    \includegraphics[width=\textwidth]{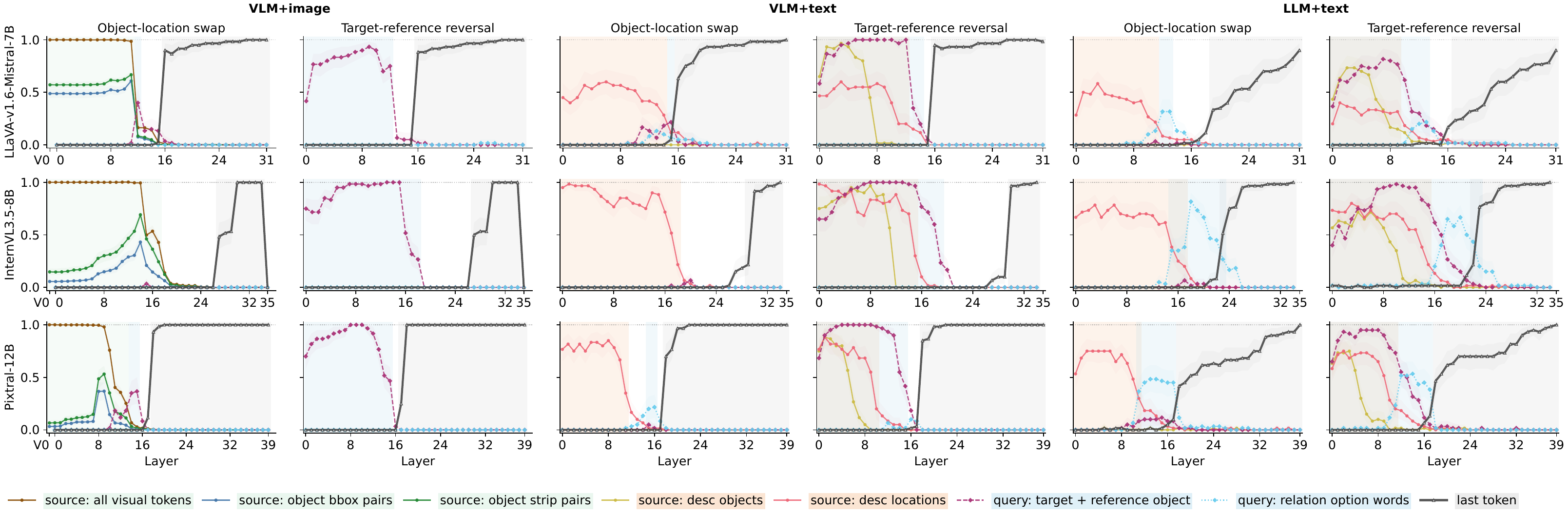}
    \caption{
    Layer-wise activation patching clean-answer recovery rate across input conditions, token sources, corruption types and models in 2-object synthetic dataset.
    }
    \label{fig:2obj-token-source-restoration}
\end{figure}

In \autoref{fig:2obj-token-source-restoration},
rows correspond to models and columns group the three input settings, VLM+image, VLM+text, and LLM+text, under two counterfactual corruptions: object-location swap and target-reference reversal. The x-axis denotes the patched layer, with V0 indicating the multimodal state before the language model. For each layer and token group, we patch clean residual activations into the corrupted run and report the fraction of examples for which the clean-answer margin is restored. Visual source tokens dominate recovery for VLM+image under object-location swaps, whereas the joint target-reference query span is most effective under target-reference reversal. In text-based settings, recovery is distributed across description object/location spans and query object spans, while late-layer final-token patching recovers the answer across models.

\begin{figure}[t]
    \centering
    \includegraphics[width=\textwidth]{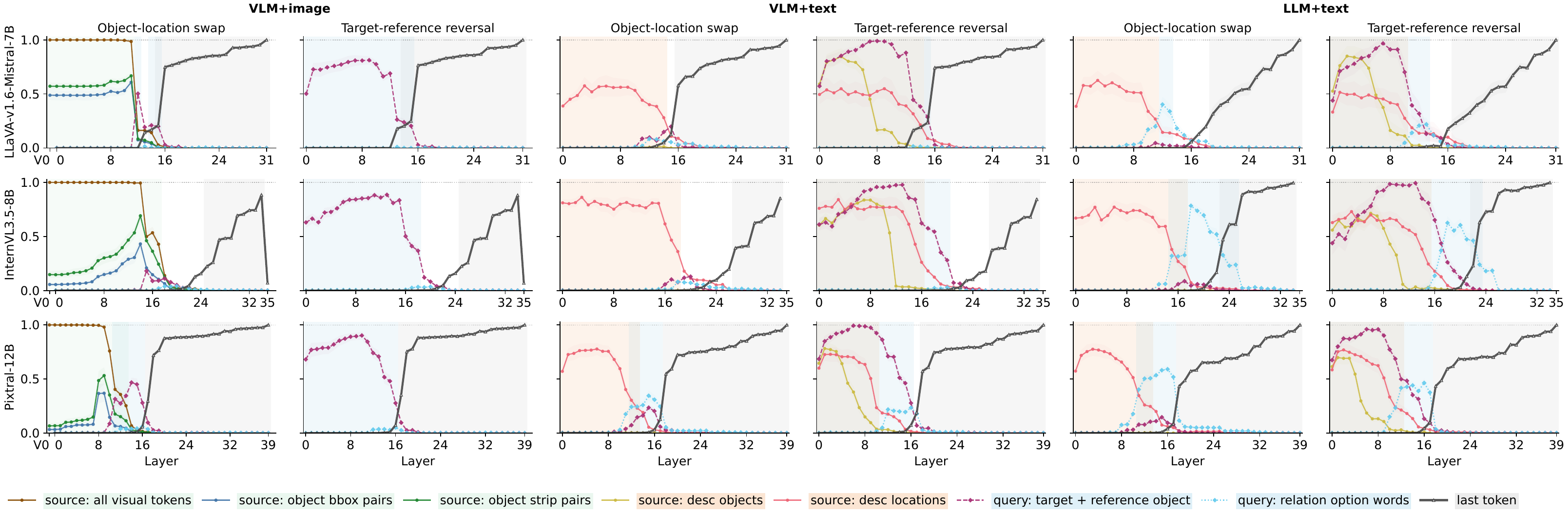}
    \caption{
    Layer-wise activation patching continuous restore score across input conditions, token sources, corruption types and models in 2-object synthetic dataset.
    }
    \label{fig:2obj-token-source-restore-score}
\end{figure}

The intervention setup in \autoref{fig:2obj-token-source-restore-score} is the same as in Figure~\ref{fig:2obj-token-source-restoration}, but the y-axis reports normalized recovery of the clean-over-corrupt answer margin. Values closer to one indicate stronger recovery toward the clean computation. The score-based results mirror the restoration-rate patterns: visual source representations provide the strongest causal signal for VLM+image under object-location swaps, joint target-reference query states are critical for target-reference reversal, and late final-token states encode the downstream answer decision.

\begin{figure}[t]
    \centering
    \includegraphics[width=\textwidth]{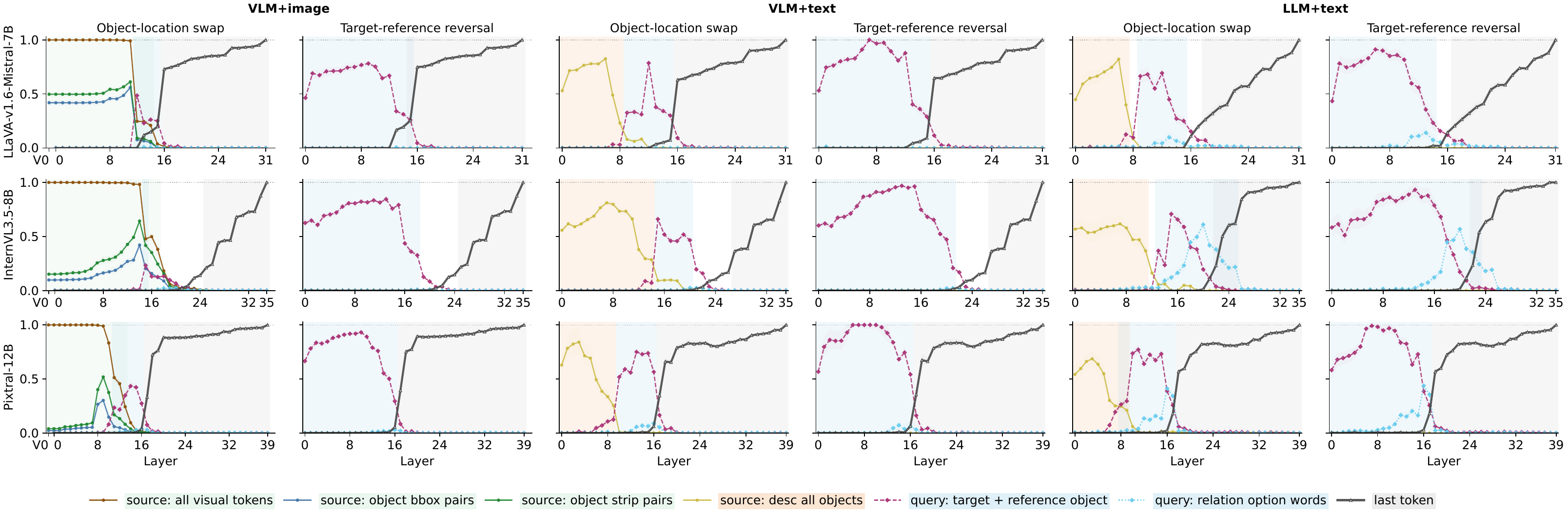}
    \caption{
    Layer-wise activation patching continuous restore score across input conditions, token sources, corruption types and models in 3-object synthetic dataset.
    }
    \label{fig:3obj-token-source-restore-score}
\end{figure}

\autoref{fig:3obj-token-source-restore-score} uses the same interventions as in Figure~\ref{fig:3obj-token-source-restoration} to measure normalized recovery of the clean answer margin after patching each token group at each layer. The margin-based results confirm that distractors do not remove the main causal structure: source-side visual tokens dominate VLM+image recovery for object-location swaps, query-side target-reference representations drive recovery for target-reference reversal, and late final-token states recover the final answer decision across models and input settings.

\section{Location ID validation results}
\label{sec:append-spatial-id-validation}

\begin{figure}[t]
    \centering
    \includegraphics[width=\linewidth]{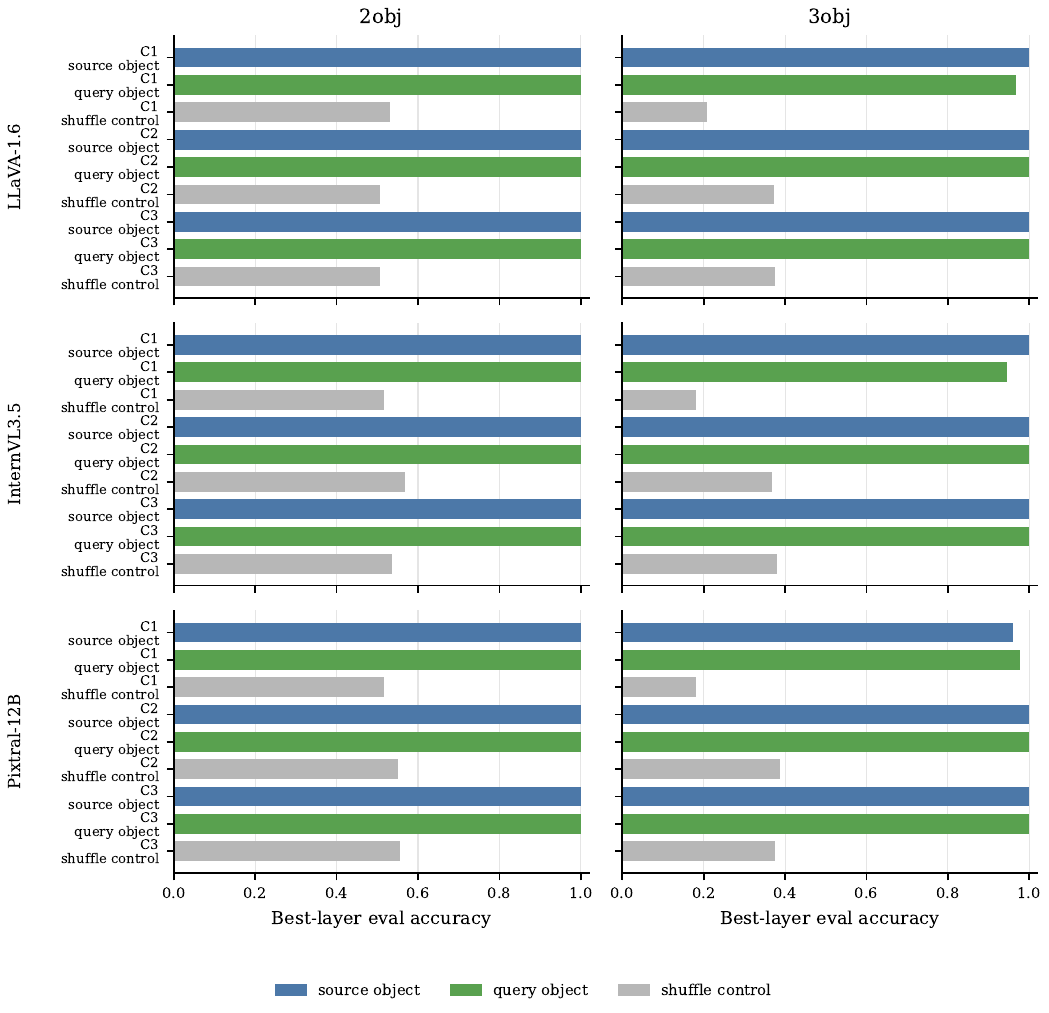}
    \caption{
    Held-out recovery and shuffle controls for extracted location IDs.
    }
    \label{fig:spatial-id-control}
\end{figure}

We summarize the held-out recovery of extracted location IDs across three models and three input conditions in \autoref{fig:spatial-id-control}. For both 2-object and 3-object location IDs, source-side and query-side IDs (\(g=\mathrm{src}\) and \(g=\mathrm{qry}\), labeled ``source object'' and ``query object'' in the figure) yield high best-layer recovery across LLaVA-1.6, InternVL3.5 and Pixtral-12B, while label-shuffled controls remain substantially lower.
This supports that the extracted IDs reflect genuine object-position binding rather than artifacts of object identity or token statistics.

\begin{figure}[tbp]
    \centering
    \begin{subfigure}{\linewidth}
        \centering
        \includegraphics[width=\linewidth]{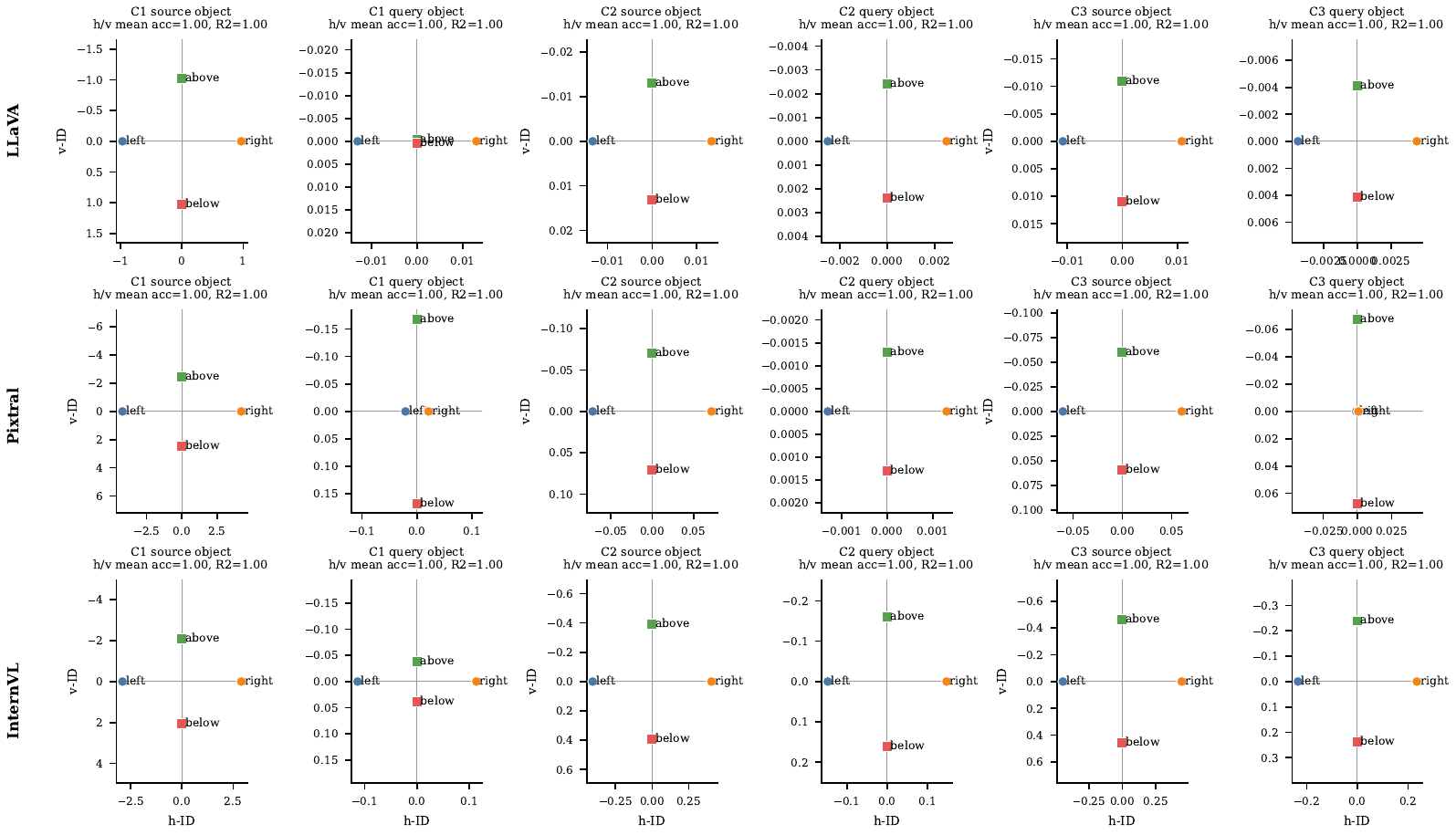}
        \caption{2-object location IDs.}
        \label{fig:spatial-id-geometry-2obj}
    \end{subfigure}

    \vspace{0.6em}

    \begin{subfigure}{\linewidth}
        \centering
        \includegraphics[width=\linewidth]{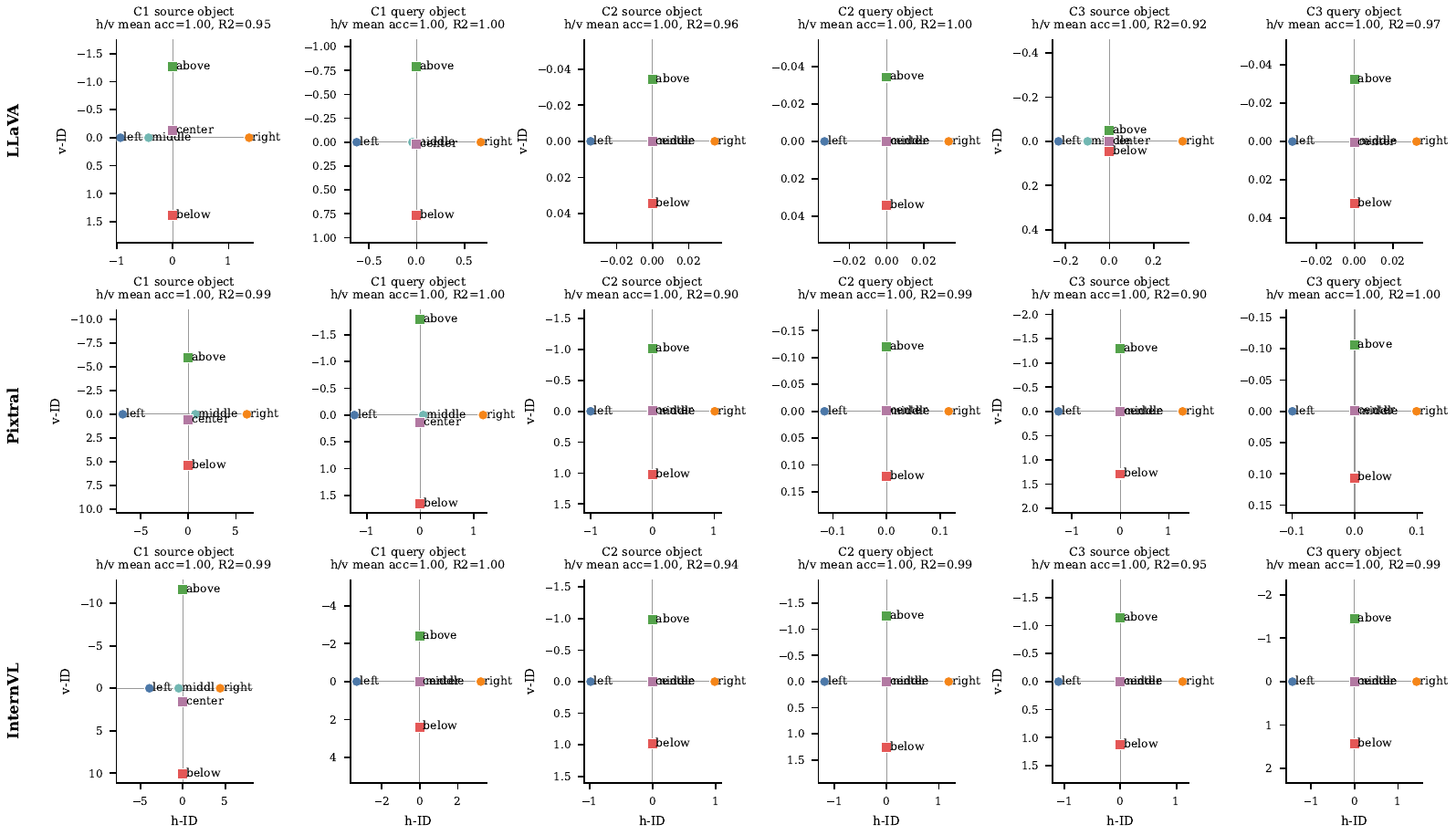}
        \caption{3-object location IDs.}
        \label{fig:spatial-id-geometry-3obj}
    \end{subfigure}

    \caption{
    Location-ID prototypes projected onto axes defined by endpoint-ID contrasts. Three-object plots additionally show the positions of middle-location prototypes relative to the endpoints.
    }
    \label{fig:spatial-id-projection-geometry}
\end{figure}

We visualize the geometry of the extracted location ID prototypes by projecting them onto the spatial axes induced by the IDs themselves. In the 2-object location-ID setting, left/right and above/below prototypes separate along the corresponding horizontal and vertical directions across all three models. In the 3-object location-ID setting, the prototypes form an ordered geometry: left/middle/right and above/center/below are arranged consistently with the ordering of their underlying spatial positions. These results in \autoref{fig:spatial-id-projection-geometry} show that the extracted IDs are not only decodable by a classifier, but also organized in a geometrically meaningful space.

\begin{figure}[tbp]
    \centering
    \includegraphics[width=\linewidth]{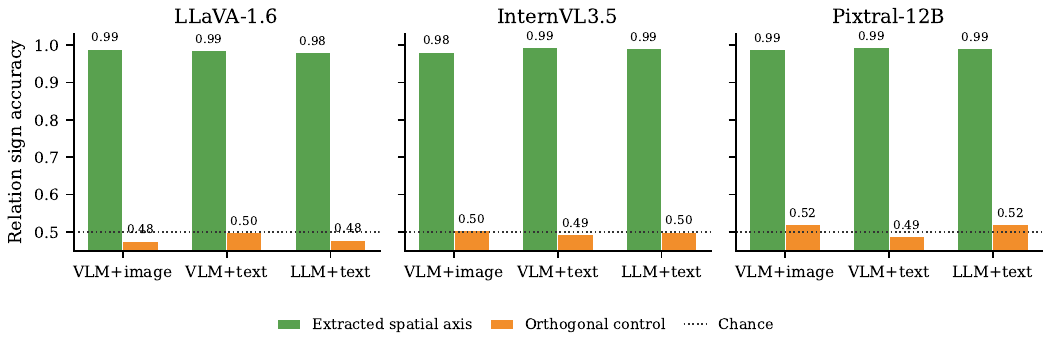}
    \caption{Query-side location information supports relational comparison across models. For each model and input setting, we project the difference between the two query-object states onto the extracted query-side spatial axis and evaluate whether the projection sign predicts their relative direction. Extracted axes achieve high relation sign accuracy, whereas matched orthogonal control axes remain close to chance.
    }
    \label{fig:query-object-spatial-axis-comparison}
\end{figure}

We further test whether location information at query-object mentions supports relational comparison.
For each model and input setting, we project the difference between the two query-object states onto the extracted query-side spatial axis \(v^{\mathrm{qry}}\) and use the sign of this projection to predict their relative direction. As shown in \autoref{fig:query-object-spatial-axis-comparison}, the extracted axes achieve consistently high relation sign accuracy under VLM+image, VLM+text and LLM+text settings, while matched orthogonal control axes remain close to chance. This suggests that the extracted IDs are not merely decodable position labels, but define axes along which object-level location information is directly comparable.

\section{Role direction intervention results}
\begin{figure}[t]
    \centering
    \includegraphics[width=\textwidth]{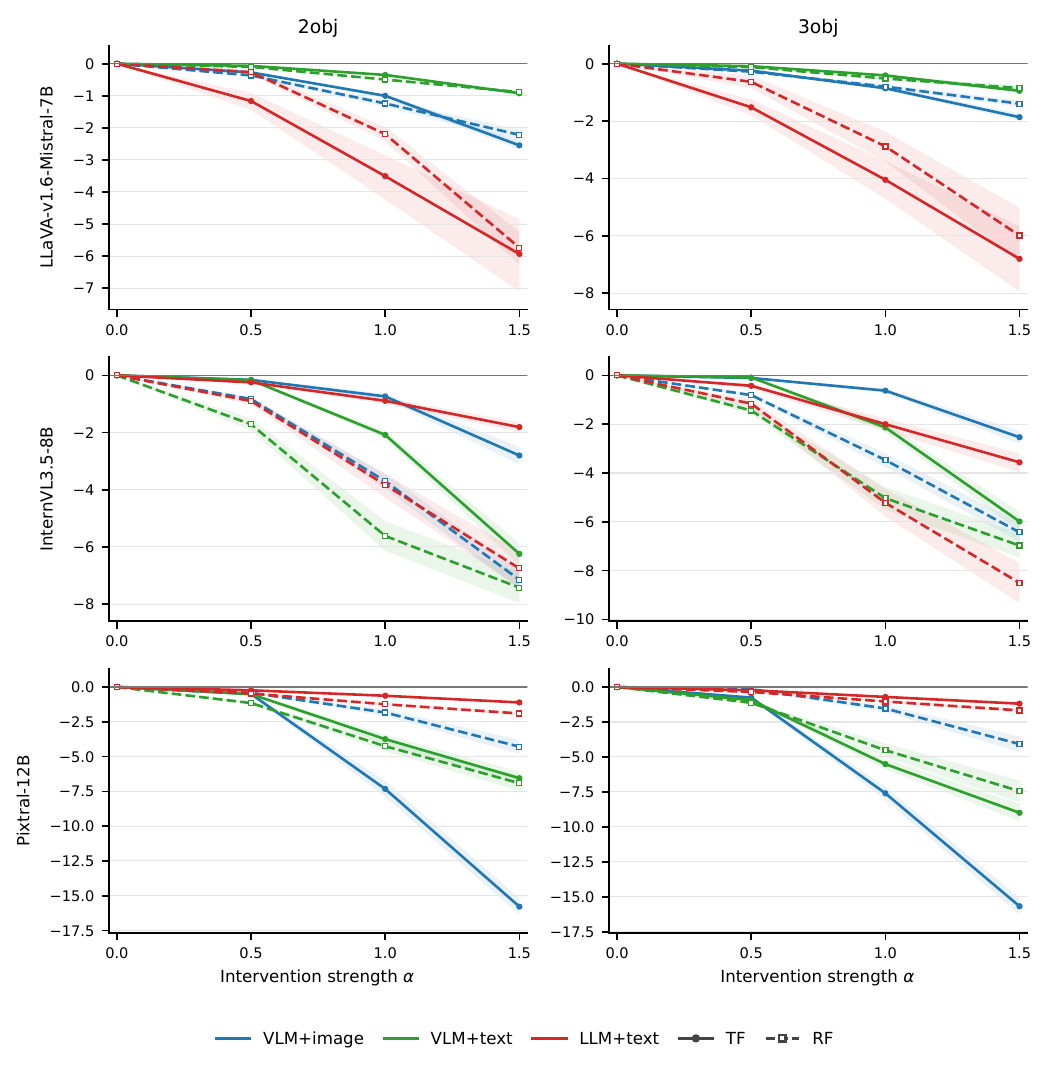}
    \caption{
    Intervention-strength curves in the selected layer bands. The y-axis uses the same direction-specific margin gap as in \autoref{fig:role_direction_layers}.
    }
    \label{fig:role_direction_dose}
\end{figure}

\begin{table}[t]
\centering
\small
\caption{Activation steering on synthetic held-out data using the joint role direction $r=(d_{\mathrm{TF}}+d_{\mathrm{RF}})/2$ estimated on synthetic training data. }
\label{tab:synthetic_steered_generation}
\resizebox{\textwidth}{!}{%
\begin{tabular}{lllccc}
\toprule
Data & Model & Condition & Accuracy &
\makecell{Target-Reference\\Reversal Consistency} &
\makecell{Location-Swap\\Consistency} \\
\midrule
2obj & LLaVA & VLM+image & $99.8 \rightarrow 100.0$ {\scriptsize $(+0.2)$} & $99.6 \rightarrow 100.0$ {\scriptsize $(+0.4)$} & $99.4 \rightarrow 100.0$ {\scriptsize $(+0.6)$} \\
2obj & LLaVA & VLM+text & $58.6 \rightarrow 60.4$ {\scriptsize $(+1.9)$} & $42.0 \rightarrow 46.3$ {\scriptsize $(+4.3)$} & $40.8 \rightarrow 45.4$ {\scriptsize $(+4.6)$} \\
2obj & LLaVA & LLM+text & $55.3 \rightarrow 59.8$ {\scriptsize $(+4.5)$} & $36.5 \rightarrow 38.1$ {\scriptsize $(+1.6)$} & $31.3 \rightarrow 34.4$ {\scriptsize $(+3.1)$} \\
\addlinespace[1pt]
2obj & InternVL & VLM+image & $100.0 \rightarrow 100.0$ {\scriptsize $(+0.0)$} & $100.0 \rightarrow 100.0$ {\scriptsize $(+0.0)$} & $100.0 \rightarrow 100.0$ {\scriptsize $(+0.0)$} \\
2obj & InternVL & VLM+text & $90.1 \rightarrow 93.3$ {\scriptsize $(+3.1)$} & $85.2 \rightarrow 88.9$ {\scriptsize $(+3.7)$} & $81.1 \rightarrow 85.5$ {\scriptsize $(+4.5)$} \\
2obj & InternVL & LLM+text & $76.1 \rightarrow 77.4$ {\scriptsize $(+1.4)$} & $52.7 \rightarrow 55.9$ {\scriptsize $(+3.1)$} & $50.6 \rightarrow 52.9$ {\scriptsize $(+2.3)$} \\
\addlinespace[1pt]
2obj & Pixtral & VLM+image & $100.0 \rightarrow 100.0$ {\scriptsize $(+0.0)$} & $100.0 \rightarrow 100.0$ {\scriptsize $(+0.0)$} & $100.0 \rightarrow 100.0$ {\scriptsize $(+0.0)$} \\
2obj & Pixtral & VLM+text & $90.0 \rightarrow 90.4$ {\scriptsize $(+0.4)$} & $80.3 \rightarrow 82.4$ {\scriptsize $(+2.1)$} & $78.9 \rightarrow 80.6$ {\scriptsize $(+1.7)$} \\
2obj & Pixtral & LLM+text & $82.5 \rightarrow 85.8$ {\scriptsize $(+3.3)$} & $65.0 \rightarrow 71.7$ {\scriptsize $(+6.6)$} & $65.5 \rightarrow 69.8$ {\scriptsize $(+4.3)$} \\
\midrule
3obj & LLaVA & VLM+image & $98.0 \rightarrow 99.0$ {\scriptsize $(+1.0)$} & $96.1 \rightarrow 98.0$ {\scriptsize $(+2.0)$} & $93.6 \rightarrow 96.1$ {\scriptsize $(+2.5)$} \\
3obj & LLaVA & VLM+text & $73.6 \rightarrow 75.8$ {\scriptsize $(+2.1)$} & $60.5 \rightarrow 63.9$ {\scriptsize $(+3.3)$} & $57.2 \rightarrow 60.7$ {\scriptsize $(+3.5)$} \\
3obj & LLaVA & LLM+text & $55.0 \rightarrow 63.5$ {\scriptsize $(+8.5)$} & $32.0 \rightarrow 39.5$ {\scriptsize $(+7.4)$} & $28.6 \rightarrow 32.9$ {\scriptsize $(+4.3)$} \\
\addlinespace[1pt]
3obj & InternVL & VLM+image & $99.8 \rightarrow 100.0$ {\scriptsize $(+0.2)$} & $99.6 \rightarrow 100.0$ {\scriptsize $(+0.4)$} & $99.8 \rightarrow 100.0$ {\scriptsize $(+0.2)$} \\
3obj & InternVL & VLM+text & $96.7 \rightarrow 97.0$ {\scriptsize $(+0.3)$} & $93.8 \rightarrow 94.1$ {\scriptsize $(+0.4)$} & $93.1 \rightarrow 94.7$ {\scriptsize $(+1.7)$} \\
3obj & InternVL & LLM+text & $83.1 \rightarrow 82.6$ {\scriptsize $(-0.5)$} & $66.6 \rightarrow 66.0$ {\scriptsize $(-0.6)$} & $62.7 \rightarrow 61.9$ {\scriptsize $(-0.8)$} \\
\addlinespace[1pt]
3obj & Pixtral & VLM+image & $100.0 \rightarrow 100.0$ {\scriptsize $(+0.0)$} & $100.0 \rightarrow 100.0$ {\scriptsize $(+0.0)$} & $100.0 \rightarrow 100.0$ {\scriptsize $(+0.0)$} \\
3obj & Pixtral & VLM+text & $88.1 \rightarrow 90.1$ {\scriptsize $(+2.1)$} & $78.1 \rightarrow 82.0$ {\scriptsize $(+3.9)$} & $76.8 \rightarrow 81.4$ {\scriptsize $(+4.7)$} \\
3obj & Pixtral & LLM+text & $72.6 \rightarrow 76.6$ {\scriptsize $(+4.0)$} & $45.1 \rightarrow 53.1$ {\scriptsize $(+8.0)$} & $44.2 \rightarrow 52.1$ {\scriptsize $(+7.8)$} \\
\bottomrule
\end{tabular}
}
\end{table}

\paragraph{Intervention-strength curves.}
To complement the layer-wise role-direction intervention results in the main text, we report intervention-strength curves within the selected layer bands. As shown in \autoref{fig:role_direction_dose}, increasing the steering strength ($\alpha$) generally makes the role-specific margin gap more negative relative to the orthogonal control. This strength-dependent effect supports the claim that the extracted target/reference role direction is causally involved in role-sensitive answer selection.

\begin{table}[t]
\centering
\caption{
Intervention configurations estimated on synthetic 2-object data for What'sUp and COCO-spatial steering. Each cell represents selected layer number/$\alpha$ value.}
\label{tab:steering_layer_alpha}
\begin{tabular}{cccc}
\toprule
Model & VLM+image & VLM+text & LLM+text \\
\midrule
LLaVA-1.6   & $10 / 1.5$ & $10 / 1.5$ & $6 / 1.5$  \\
InternVL3.5 & $12 / 1.0$ & $15 / 1.5$ & $14 / 1.0$ \\
Pixtral     & $9 / 1.5$  & $8 / 1.5$  & $6 / 1.0$  \\
\bottomrule
\end{tabular}
\end{table}

\paragraph{Steering configurations.}
\autoref{tab:steering_layer_alpha} lists the intervention
layers and strengths used for What'sUp and COCO-spatial evaluation.
Layers follow the previously selected synthetic-validation
policy, while strengths are selected on separate synthetic
two-object validation data.

\paragraph{Synthetic held-out steering results.}
We also evaluate whether role-direction steering improves prediction on synthetic held-out examples, using directions and hyperparameters selected only on synthetic validation data. \autoref{tab:synthetic_steered_generation} shows that steering often improves target-reference reversal and location-swap consistency, especially in text-based conditions where the baseline is lower and there is more room for improvement. In contrast, VLM+image results are frequently saturated, leaving little room for additional gains. These results provide an in-domain counterpart to the What'sUp transfer results in the main text and show that the same role-direction intervention can improve paired consistency without updating model parameters.

\paragraph{COCO-spatial steering results.}
To evaluate the transfer of role-direction steering to natural images, we conduct experiments on the two-object subset of COCO-Spatial, comprising 440 annotated object pairs across 295 images. For each model, we reuse the joint target/reference role direction, intervention layer, and steering strength determined on synthetic data, without retuning them on COCO-Spatial. We compare baseline and steered responses using accuracy on the original query and target-reference reversal consistency, which requires both the original and reversed questions to be answered correctly. The results are displayed in \autoref{tab:steered_generation_coco_spatial}.

\begin{table}[t]
\centering
\small
\caption{Activation steering results of the joint target/reference role direction on COCO-Spatial. Bold changes indicate improvements with 95\% confidence intervals excluding zero. Location-swap consistency is unavailable because the evaluated subset does not contain paired original and location-swapped scenes.}
\label{tab:steered_generation_coco_spatial}
\resizebox{\textwidth}{!}{%
\begin{tabular}{ccccc}
\toprule
Model & Condition & Accuracy &
\makecell{Target-Reference\\Reversal Consistency} &
\makecell{Location-Swap\\Consistency} \\
\midrule
LLaVA-1.6 & VLM+image & $93.2 \rightarrow 94.5$ {\scriptsize $(+1.4)$} & $87.7 \rightarrow 91.1$ {\scriptsize $\mathbf{(+3.4)}$} & N/A \\
LLaVA-1.6 & VLM+text & $69.9 \rightarrow 69.2$ {\scriptsize $(-0.7)$} & $45.1 \rightarrow 48.0$ {\scriptsize $\mathbf{(+2.8)}$} & N/A \\
LLaVA-1.6 & LLM+text & $64.0 \rightarrow 66.9$ {\scriptsize $\mathbf{(+3.0)}$} & $36.9 \rightarrow 39.2$ {\scriptsize $(+2.3)$} & N/A \\
\midrule
InternVL3.5 & VLM+image & $97.0 \rightarrow 97.0$ {\scriptsize $(+0.0)$} & $93.9 \rightarrow 94.8$ {\scriptsize $\mathbf{(+0.9)}$} & N/A \\
InternVL3.5 & VLM+text & $88.1 \rightarrow 93.0$ {\scriptsize $\mathbf{(+4.9)}$} & $82.0 \rightarrow 89.4$ {\scriptsize $\mathbf{(+7.4)}$} & N/A \\
InternVL3.5 & LLM+text & $71.5 \rightarrow 76.5$ {\scriptsize $\mathbf{(+5.0)}$} & $49.2 \rightarrow 62.4$ {\scriptsize $\mathbf{(+13.2)}$} & N/A \\
\midrule
Pixtral & VLM+image & $90.2 \rightarrow 91.8$ {\scriptsize $(+1.6)$} & $81.4 \rightarrow 83.6$ {\scriptsize $(+2.3)$} & N/A \\
Pixtral & VLM+text & $85.2 \rightarrow 85.1$ {\scriptsize $(-0.1)$} & $68.4 \rightarrow 71.7$ {\scriptsize $\mathbf{(+3.3)}$} & N/A \\
Pixtral & LLM+text & $76.3 \rightarrow 78.2$ {\scriptsize $\mathbf{(+1.9)}$} & $49.5 \rightarrow 52.8$ {\scriptsize $\mathbf{(+3.3)}$} & N/A \\
\bottomrule
\end{tabular}
}
\end{table}

\paragraph{GQA-spatial steering results.}
We further evaluate the transfer of role-direction steering on the left/right subset of the two-object GQA-Spatial benchmark, comprising 264 spatial-relation annotations across 233 images. For each model, we reuse the joint target/reference role direction, intervention layer, and steering strength determined on synthetic data, without retuning them on GQA-Spatial. We compare baseline and steered responses using accuracy on the original query and target-reference reversal consistency, which requires both the original and reversed questions to be answered correctly. The results are displayed in \autoref{tab:steered_generation_gqa_spatial}.

\begin{table}[t]
\centering
\small
\caption{Activation steering results of the joint target/reference role direction on the left/right subset of GQA-Spatial. Bold changes indicate improvements with 95\% confidence intervals excluding zero.}
\label{tab:steered_generation_gqa_spatial}
\resizebox{\textwidth}{!}{%
\begin{tabular}{ccccc}
\toprule
Model & Condition & Accuracy &
\makecell{Target-Reference\\Reversal Consistency} &
\makecell{Location-Swap\\Consistency} \\
\midrule
LLaVA-1.6 & VLM+image & $93.6 \rightarrow 95.8$ {\scriptsize $\mathbf{(+2.3)}$} & $90.9 \rightarrow 93.9$ {\scriptsize $\mathbf{(+3.0)}$} & N/A \\
LLaVA-1.6 & VLM+text & $57.6 \rightarrow 58.0$ {\scriptsize $(+0.4)$} & $49.8 \rightarrow 49.6$ {\scriptsize $(-0.2)$} & N/A \\
LLaVA-1.6 & LLM+text & $44.5 \rightarrow 48.5$ {\scriptsize $\mathbf{(+4.0)}$} & $13.1 \rightarrow 19.1$ {\scriptsize $\mathbf{(+6.1)}$} & N/A \\
\midrule
InternVL3.5 & VLM+image & $98.1 \rightarrow 98.5$ {\scriptsize $(+0.4)$} & $97.7 \rightarrow 97.7$ {\scriptsize $(+0.0)$} & N/A \\
InternVL3.5 & VLM+text & $81.6 \rightarrow 88.1$ {\scriptsize $\mathbf{(+6.4)}$} & $71.8 \rightarrow 83.1$ {\scriptsize $\mathbf{(+11.4)}$} & N/A \\
InternVL3.5 & LLM+text & $72.2 \rightarrow 77.7$ {\scriptsize $\mathbf{(+5.5)}$} & $48.3 \rightarrow 66.1$ {\scriptsize $\mathbf{(+17.8)}$} & N/A \\
\midrule
Pixtral & VLM+image & $95.8 \rightarrow 96.2$ {\scriptsize $(+0.4)$} & $92.8 \rightarrow 93.9$ {\scriptsize $(+1.1)$} & N/A \\
Pixtral & VLM+text & $75.6 \rightarrow 78.0$ {\scriptsize $\mathbf{(+2.5)}$} & $53.4 \rightarrow 59.8$ {\scriptsize $\mathbf{(+6.4)}$} & N/A \\
Pixtral & LLM+text & $58.0 \rightarrow 60.2$ {\scriptsize $\mathbf{(+2.3)}$} & $21.8 \rightarrow 25.6$ {\scriptsize $\mathbf{(+3.8)}$} & N/A \\
\bottomrule
\end{tabular}
}
\end{table}

\clearpage
\section{Implementation Details and Compute Resources}
\label{app:compute}

\paragraph{Hardware.}
Model inference and GPU-based interventions were run on a shared cluster using one single NVIDIA A100-SXM4 GPU with 40\,GB of
device memory. 
Analyses operating on cached activations, including
location-ID estimation, were also scheduled as separate CPU jobs.

\paragraph{Software and intervention implementation.}
The evaluation environment used Python 3.13.1, PyTorch 2.9.1
with CUDA 12.8, Hugging Face Transformers 4.57.6, and
Accelerate 1.12.0.
Activation extraction, activation patching, location-ID interventions,
and role-direction steering were implemented using custom PyTorch
forward and forward-pre hooks.
During autoregressive generation, interventions on prompt-token
representations were applied during prefill.
Pretrained model parameters remained frozen throughout these experiments.

\paragraph{Numerical precision.}
Models were evaluated without weight quantization.
LLaVA models, Vicuna backbones, and Mistral-7B used FP16 in the
aligned evaluation and intervention pipeline.
Pixtral-12B, Mistral-Nemo-12B, InternVL3.5, and the Qwen3/Qwen3-VL
models used BF16.

\paragraph{Inputs and decoding.}
We used greedy decoding with sampling disabled.
The aligned behavioral benchmark and the main constructive
role-direction steering evaluations used a maximum of eight new tokens.
In the VLM+text condition, the model received a textual scene
description together with a blank white image.
Generated-answer correctness was evaluated from the decoded response.
Candidate-answer scores were evaluated separately: single-token labels
were scored at the final prompt position, while multi-token labels,
where required, were scored using teacher-forced sequence
log-likelihoods.

\paragraph{Random-direction controls.}
For the main role-direction experiments, random-control results were
averaged over five seeds, $\{42,43,44,45,46\}$.
Control directions were orthogonal to the corresponding role direction
and matched in norm, using the same intervention sites and strengths.
The seeds controlled the construction of random directions, and
answer generation remained greedy.

\paragraph{Uncertainty estimates.}
We estimate pointwise 95\% confidence intervals for
steered-minus-baseline metric differences using a paired cluster
bootstrap with 2,000 replicates. We use random seed 20260816 for What'sUp and 20260821 for COCO-Spatial and GQA-Spatial.
The resampling unit is the scene family for What'sUp and the original image for COCO-Spatial and GQA-Spatial. 
Queries and variants within each unit share the same resampling multiplicity,
with baseline and steered outcomes kept paired. 
Each replicate preserves the original metric definitions and aggregation weights.
Intervals are defined by the 2.5th and 97.5th percentiles of the resulting differences. 

\end{document}